\def\cvprPaperID{3812}
\def\confName{CVPR}
\def\confYear{2023}

\def\paperTitle{StyleGAN Salon:\\ Multi-View Latent
Optimization for Pose-Invariant Hairstyle Transfer}

\def\authorBlock{
    Sasikarn Khwanmuang$\vcenter{\hbox{\includegraphics[width=0.013\textwidth\vspace*{0.3em}]{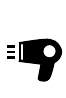}}}$  \quad 
    Pakkapon Phongthawee$\vcenter{\hbox{\includegraphics[width=0.013\textwidth\vspace*{0.3em}]{figs/unicode_barber.pdf}}}$  \quad
    Patsorn Sangkloy$\vcenter{\hbox{\includegraphics[width=0.012\textwidth\vspace*{0.3em}]{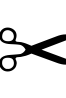}}}$ \quad 
    Supasorn Suwajanakorn$\vcenter{\hbox{\includegraphics[width=0.013\textwidth\vspace*{0.3em}]{figs/unicode_barber.pdf}}}$ \\
    $\vcenter{\hbox{\includegraphics[width=0.013\textwidth\vspace*{0.4em}]{figs/unicode_barber.pdf}}}$VISTEC, Thailand   \qquad $\vcenter{\hbox{\includegraphics[width=0.012\textwidth\vspace*{0.4em}]{figs/unicode_scissors.pdf}}}$Phranakhon Rajabhat University, Thailand

 \\
    {\tt\small \{sasikarn.k\_s18, pakkapon.p\_s19, supasorn.s\}@vistec.ac.th \qquad patsorn.s@pnru.ac.th}

}

\newif\ifreview 
\newif\ifarxiv 
\newif\ifcamera \newcommand{\cameraready}{\cameratrue}
\newif\ifrebuttal

\cameraready

\pdfoutput=1
\documentclass[10pt,twocolumn,letterpaper]{article}
\ifreview \usepackage[review]{cvpr} \fi
\ifarxiv \usepackage[pagenumbers]{cvpr} \fi
\ifrebuttal \usepackage[rebuttal]{cvpr} \fi
\ifcamera \usepackage{cvpr} \fi

\usepackage{graphicx}
\usepackage{amsmath}
\usepackage{amssymb}
\usepackage{booktabs}

\usepackage{times}
\usepackage{microtype}
\usepackage{epsfig}
\usepackage[table,xcdraw]{xcolor}
\usepackage{caption}
\usepackage{float}
\usepackage{placeins}
\usepackage{color, colortbl}
\usepackage{stfloats}
\usepackage{paralist}
\usepackage{enumitem}
\usepackage{tabularx}
\usepackage{xstring}
\usepackage{multirow}
\usepackage{xspace}
\usepackage{url}
\usepackage{subcaption}
\usepackage{xcolor}
\usepackage[hang,flushmargin]{footmisc}

\usepackage[T1]{fontenc}  
\usepackage[utf8]{inputenc}
\DeclareUnicodeCharacter{2702}{\ensuremath{\scissor}}

\usepackage[accsupp]{axessibility}  

\usepackage{booktabs} 
\usepackage{amsmath,bm}
\usepackage{pifont}
\newcommand{\cmark}{\ding{51}}%
\usepackage[table,xcdraw]{xcolor}

\usepackage{xcolor}

\usepackage[normalem]{ulem}

\newcommand{\iguide}{I_\text{guide}}
\newcommand{\iguidef}{I_\text{guide}^\text{face}}
\newcommand{\iguideh}{I_\text{guide}^\text{hair}}

\newcommand{\W}{\mathcal{W}}
\newcommand{\WP}{\mathcal{W^+}}
\newcommand{\FS}{\mathcal{F/S}}

\ifcamera \usepackage[accsupp]{axessibility} \fi

% ...

\ifarxiv  \fi

\definecolor{BrickRed}{RGB}{196,49,25}
\definecolor{ForestGreen}{RGB}{34,139,34}
\definecolor{BlueViolet}{RGB}{45,52,151}

\newcommand{\R}[1]{{%
    \textbf{%
        \ifstrequal{#1}{1}{\textcolor{BrickRed}{R#1}}{
        \ifstrequal{#1}{2}{\textcolor{BlueViolet}{R#1}}{
        \ifstrequal{#1}{3}{\textcolor{magenta}{R#1}}{%
        \ifstrequal{#1}{4}{\textcolor{ForestGreen}{R#1}}{
                           \textcolor{cyan}{R#1}%
        }}}}%
    }%
}}

\usepackage{xr-hyper}

\makeatletter
\newcommand*{\addFileDependency}[1]{
  \typeout{(#1)}
  \@addtofilelist{#1}
  \IfFileExists{#1}{}{\typeout{No file #1.}}
}

\makeatother

\usepackage[pagebackref,breaklinks,colorlinks]{hyperref}
\usepackage[capitalize]{cleveref}
\crefname{section}{Sec.}{Secs.}
\crefname{table}{Table}{Tables}
\crefname{figure}{Fig.}{Figs.}

\usepackage{amsmath}
\begin{document}
\title{\paperTitle}
\author{\authorBlock}
\twocolumn[{%
\renewcommand\twocolumn[1][]{#1}%
\maketitle
\begin{center}
\centering
  \includegraphics[width=1\linewidth\vspace*{-0.4em}]{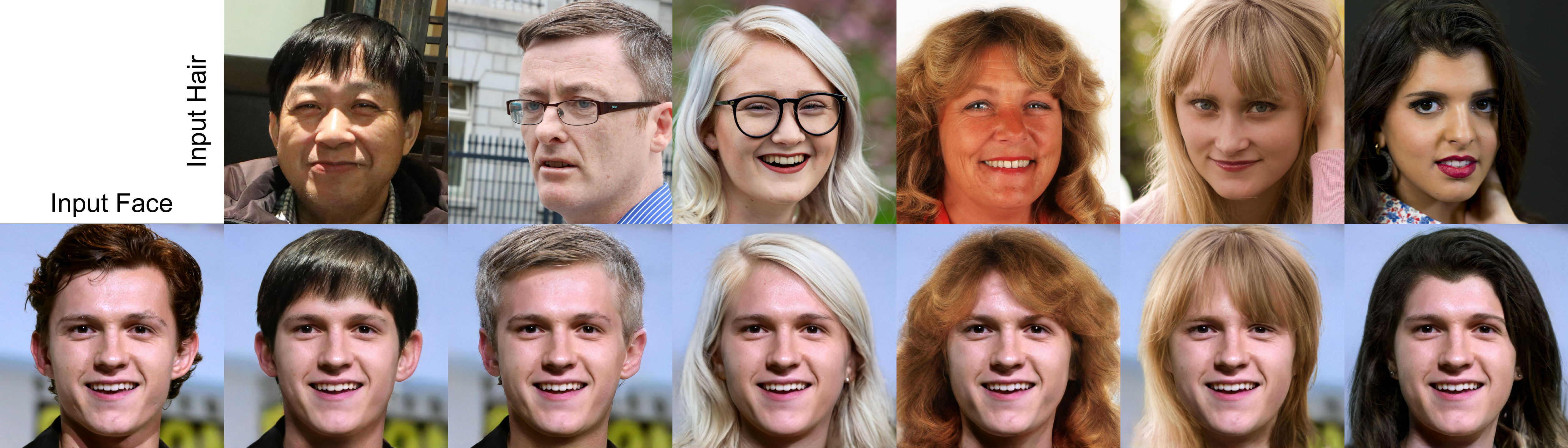}
  \captionof{figure}{Our method can transfer the hairstyle from any reference hair image in the top row to Tom Holland~\cite{tom}, in the second row. }
  \label{fig:teasor}
\end{center}
}]

\begin{abstract}
Our paper seeks to transfer the hairstyle of a reference image to an input photo for virtual hair try-on. 
We target a variety of challenges scenarios, 
such as transforming a long hairstyle with bangs to a pixie cut, which requires removing the existing hair and inferring how the forehead would look, or transferring partially visible hair from a hat-wearing person in a different pose.
Past solutions leverage StyleGAN for hallucinating any missing parts and producing a seamless face-hair composite through so-called GAN inversion or projection. 
However, there remains a challenge in controlling the hallucinations to accurately transfer hairstyle and preserve the face shape and identity of the input.
To overcome this, we propose a multi-view optimization framework that uses \emph{two different views} of reference composites to semantically guide occluded or ambiguous regions. Our optimization shares information between two poses, which allows us to produce high fidelity and realistic results from incomplete references.
Our framework produces high-quality results and outperforms prior work in a user study that consists of significantly more challenging hair transfer scenarios than previously studied. 
Project page: \emph{\small \url{https://stylegan-salon.github.io/}}.
\end{abstract}
\section{Introduction}
\label{sec:intro}
What makes Jennifer Aniston keep her same hairstyle for over three decades? Perhaps she likes the classic, or perhaps changing her hairstyle is a decision too high-stakes that she could later regret. Unlike garments or makeup, trying on a new hairstyle is not easy, and being able to imagine yourself in different hairstyles could be an indispensable tool.

Recent approaches for hairstyle transfer, 
StyleYourHair~\cite{kim2022styleyourhair}, 
Barbershop~\cite{barbershop}, LOHO~\cite{saha2021LOHO}, and MichiGAN~\cite{tan2020michigan}, allow users to manipulate multiple hair attributes of an input image, such as appearance, shape, or color by providing \emph{a reference image} for each different attribute. These methods \cite{kim2022styleyourhair, barbershop, saha2021LOHO} rely on a generative adversarial network~\cite{goodfellow2014generative}, specifically StyleGAN2~\cite{stylegan2}, which can synthesize highly realistic face images. Their key idea, which also forms the basis of our method, is to leverage the realistic face distribution learned by StyleGAN and search for a hairstyle-transfer output whose latent code lies within the learned distribution using optimization (commonly known as GAN projection or inversion). 

Our extensive study on these state-of-the-art techniques still reveal several unsolved challenges for \emph{in-the-wild hairstyle transfer}.
One of the main challenges is when the reference hair comes from a person with a very different head pose or facial shape. In this case, the transfer result often degrades significantly ~\cite{saha2021LOHO, barbershop}.
HairFIT~\cite{chung2022hairfit} and the recently proposed StyleYourHair~\cite{kim2022styleyourhair} both attempt to solve this using an additional alignment step to align the pose of the target hair to the input face. In HairFIT~\cite{chung2022hairfit}, this is done explicitly via a flow-based warping module for hair segmentation, but this requires training on multi-view datasets~\cite{nagrani2017voxceleb,kim2021k}. StyleYourHair~\cite{kim2022styleyourhair} avoids this issue and also improves upon HairFIT's results by optimizing for an aligned pose within StyleGAN2's latent space using distances between the detected facial keypoints.
While StyleYourHair can handle a certain degree of misalignment, it often struggles to preserve details of the reference hair texture, especially for intricate and non-straight hairstyles, \eg, in Figure~\ref{fig:comparision1}.

In general, we have observed that there is a trade-off between 
hallucinating new details, which is crucial for handling highly different poses, and 
preserving the original texture from the reference images. 
These two goals are often at odds with each others. This trade-off is also evident in the results from StyleYourHair, as reported in \cite{kim2022styleyourhair} and as shown in table~\ref{table:both_recon}, where BarberShop~\cite{barbershop} can still produce better results, for example, when the pose difference is small.


We tackle this dilemma by performing a multi-stage optimization. This serves two purposes: first, to hallucinate new details necessary for aligning the poses, and second, to recover face-hair details from the original input images. We preserve these details in a form of two guide images in \emph{both} viewpoints, which will be jointly optimized to allow new details to be filled while retaining face-hair texture in the original pose. 
Our pose alignment is done both explicitly via 3D projection~\cite{eg3d} on \emph{RGB} images, and implicitly via latent code(s) sharing during our multi-view optimization. 
In summary, our contributions are as follows:
\begin{enumerate} 
\item We propose StyleGAN Salon: a pose-invariant hairstyle transfer pipeline that is flexible enough to handle a variety of challenging scenarios including, but not limited to, bangs/hat removal and background inpainting.
\item Unlike previous works, our method operates entirely on RGB images, which are more flexible than segmentation masks. This allows us to first \emph{draft} the output and then refine them via multi-stage optimization. 
\item We introduce multi-view optimization for hairstyle transfer which incorporates 3D information to align the poses for both face and hair images. Our method leverages both views to help preserve details from the original images. 
\item We thoroughly analyze the results in several experiments, including a user study with detailed breakdown into various challenging scenarios. Our method shows superior results over existing works in \emph{all} scenarios. 
\end{enumerate}

\section{Related Work}
\label{sec:related}


\indent\textbf{Generative Adversarial Networks.} Beginning with~\cite{goodfellow2014generative,stylegan}, StyleGANs~\cite{stylegan,stylegan2,stylegan3} have shown great success in 2D image generation by learning realistic training data distribution corresponding to a fixed low-dimensional distribution, called latent code. 
The learned latent code can be applied to a downstream task such as image manipulation~\cite{shen2020interfacegan,härkönen2020ganspace,he2019attgan,wu2019relgan}.
Recent work on \cite{eg3d,gu2021stylenerf,zhao-gmpi2022} 
also explore 3D aware architecture in StyleGAN, 
resulting in multi-view consistent images. 

For hairstyle transfers, MichiGAN~\cite{tan2020michigan} uses a conditional hair generation network that can control hair shape, structure, and appearance.
Recently, HairFIT~\cite{chung2022hairfit}, a pose-invariant hairstyle transfer network, aligns the reference hair to match the input face pose using a optical flow-based hair alignment module, but requires training on multi-view dataset~\cite{nagrani2017voxceleb,kim2021k}. 

\indent\textbf{StyleGAN Latent Space and Projection Techniques.}
To generate an image, StyleGAN2 first maps a random latent code $z \sim N(\mathbf{0}, I)$ to an intermediate latent code $w \in \mathbb{R}^{512}$ in a new space called $\W$. This code $w$ is then replicated (18x) and used as input to each layer of the generator that controls details at different scales through weight demodulation. Collectively, these replicated latent codes are referred to as $w^+ \in \mathbb{R}^{18 \times 512}$ in the extended latent space $\WP$. 

Pivotal Tuning Inversion (PTI)~\cite{pti} further improves the projection quality by directly tuning generator weights around the optimized latent code $w$ using their regularization technique. 
Other techniques \cite{image2stylegan,image2stylegan++,thatgithub} optimize the latent code $w^+$ in each layer separately to better match the input image.

PIE~\cite{pie} introduces hierarchical optimization for semantic attributes editing that first optimizes the latent code in the $\W$ space, then transfers the code to the $\WP$ space and continues the optimization in that space. 
We adopt similar hierarchical optimization that uses both $\W$ and $\WP$. 
Part of our method is also inspired by PULSE~\cite{pulse}, which reconstructs a high-resolution image from a small reference image (32x32) by searching for the closest image in the StyleGAN latent space that resembles the low-resolution reference.

Some methods~\cite{xu2021continuity,zhou2018multiview,chandran2021rendering} project multiple images into latent space simultaneously. However, all of their inputs are complete, whereas our method requires hair information from the reference hair image and additional information from the input face image.

\indent\textbf{StyleGAN-Based Hairstyle Transfer.}
LOHO~\cite{saha2021LOHO} adopts loss functions from Image2StyleGAN++~\cite{image2stylegan++} to combine face-hair inputs into a hairstyle transfer output with their orthogonalization optimization technique which reduces conflicts between multiple loss functions.
Barbershop~\cite{barbershop} 
first predicts semantic regions of both inputs and uses them to create a \emph{``target segmentation mask''} of the output with rule- (in the original paper) or GAN-based inpainting (in their official code). 
Then, they optimize two separate latent codes in $\WP$ space, one for matching the face and the other for hair, 
while conforming to the target segmentation mask.
To preserve the original details, the latent optimization is done in their proposed $\FS$ space, which replaces the first seven blocks of $\WP$ space with the corresponding activation maps of StyleGAN's convolution layers. 
To improve Barbershop's capacity to handle unaligned input poses, StyleYourHair~\cite{kim2022styleyourhair} first aligns the reference hair to match the pose of the input face via the proposed local-style-matching loss.
However, this alignment often leads to an unrealistic hair shape or inaccurate hair texture results. In contrast, our method simultaneously optimizes for two guides in face and hair views, resulting in a better hair texture.

Instead of estimating the final output from the segmentation mask~\cite{kim2022styleyourhair,barbershop}, our multi-view optimization uses face-hair composites in RGB space 
to overcome this problem, and produces results that better preserve the input facial structure and hairstyle across a wider, more challenging range of scenarios.
Concurrent work, HairNet~\cite{zhu2022hairnet}, arrives at a similar goal of removing the target segmentation mask via a two-step process that involves baldification using StyleFlow~\cite{abdal2021styleflow} followed by another network to transfer the hairstyle. 

Nevertheless, handling pose differences is crucial for successful hairstyle transfer; our method incorporates 3D rotation to preserve the geometric consistency and systematically evaluate this aspect.

\section{StyleGAN Salon}
\label{sec:approach}
Given two input images $I_h$ and $I_f$, the goal is to transfer the hair from the reference hair image $I_h$ into the face in the image $I_f$, while preserving all remaining details in $I_f$, including identity, clothing, and background.

The key idea of our approach is to guide the optimization on the learned latent space of StyleGAN2 with two ``guide'' images, which represent rough composites of the final output based on simple cut-and-paste in the viewpoint of $I_h$ and $I_f$. We leverage EG3D to construct these guided images in a geometrically consistent way, described in Section \ref{sec:met_align}. 

Optimization on StyleGAN2's latent space is commonly performed on either the original latent space $\W$ or the extended latent space $\WP$. 
Optimizing on $\W$ space generally leads to more realistic results by staying within the original latent space, whereas optimizing on $\WP$ allows a closer match to the reference \cite{pie}. 
In hairstyle transfer, it is important to stay faithful to the input images and preserve important details such as hair texture, face identity, and background. However, optimizing on $\WP$ will lead to poor results because our guide images are rough and unrealistic composites. Thus, we propose to optimize on $\W$ space followed by $\WP$ space, similar to a technique in PIE~\cite{pie} used for editing semantic attributes of an image. 

\begin{figure*}
\centering
\includegraphics[width=1\linewidth\vspace*{-0.4em}]{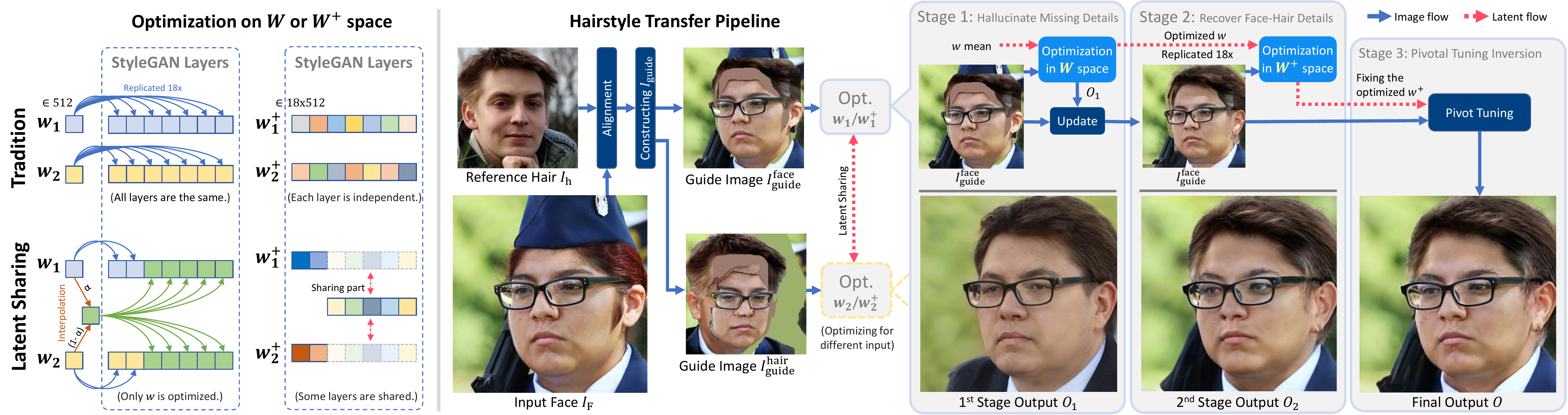}
  \caption{\textbf{Overview of StyleGAN Salon:}
  We first align the input face $I_f$ and reference hair $I_h$ and use them to construct guide images $\iguide$, in two different viewpoints, which specifies the target appearance for each output region. (see Section \ref{sec:approach})
  }
  
\label{fig:overview_pre}
\end{figure*}

Our optimization incorporates both guide images from the two viewpoints, detailed in Section \ref{sec:met_multi}.
Section \ref{sec:met_hall} and Section \ref{sec:met_details} cover details of our optimization on $\W$ and $\WP$, respectively.
Finally, we also optimize StyleGAN2 weights while freezing the latent codes 
(Section \ref{sec:met_pti}) using PTI~\cite{pti} to further improve detail fidelity. Figure \ref{fig:overview_pre} shows an overview of our complete pipeline.

\subsection{Constructing the Guide Images}\label{sec:met_align}
The purpose of our guide images is to provide an initial estimate of how the hair would look on $I_f$. We achieve this using a cut-and-paste composite of the face and background from $I_f$ and the hair from $I_h$. To better handle a potentially large shift in viewpoint between $I_f$ and $I_h$, we propose to leverage EG3D~\cite{eg3d} to help generate geometrically consistent guide images. 
We argue that using multi-view guide images, each in the pose of $I_f$ and $I_h$, helps preserve details that could otherwise be lost from using a single viewpoint alone.


A straightforward approach is to simply project $I_h$ into the EG3D~\cite{eg3d} latent space, and use their proposed neural rendering pipeline to render it in the view of $I_f$ (and vice versa). However, while the projection can produce a geometrically consistent 3D shape, we found that the resulting texture is not very accurate. To address this, we replace any visible regions in the texture from $I_h$'s viewpoint with the original pixels of $I_h$, while leaving the rest of the texture as the projected texture from EG3D's rendering.
We also additionally apply uniform scaling and translation to match the faces' widths and centers, which are computed based on detected facial keypoints~\cite{dlib09}. These keypoints, along with semantic regions from~\cite{yu2018bisenet}, are also used to handle various corner cases, such as re-painting unwanted hair regions. We refer to Appendix \ref{sec:det} for more details of this operation.

The entire process is done for both $I_f$ and $I_h$ viewpoints, resulting in a pair of guide images $\iguide = [ \iguidef, \iguideh ]. $
We emphasize that by leveraging our multi-view optimization (Section
\ref{sec:met_multi}), these guide images do not need to be precise or realistic
to produce convincing final results, as demonstrated in the first row of Figure
\ref{fig:ablation}.


\subsection{Multi-View Latent Optimization}\label{sec:met_multi}
Our guide images provide complementary information about the target hair and face, albeit from different poses. However, each guide image is only fully accurate in regions with the original pixels seen in the original viewpoint and not warped by EG3D. 
Thus, our optimization goal is to combine information from both guide images to generate a final output that accurately captures the realistic hair from $\iguideh$, as well as other details from $\iguidef$. We achieve this using multiple loss functions that attend to both viewpoints with different spatial emphasis. Specifically, we perform multi-view latent optimization on $w$/$w^+$ (and stochastic noise maps $n$) that fits both guide images:
\begin{equation}
    \min_{\{ w / w^+, n \}}  \sum_\text{i}^{[\text{face, hair}]}
     \mathcal{L}_\text{loss} (O^\text{(i)}, I_\text{guide}^\text{(i)}),
\end{equation}
where $O^\text{(i)}$ denotes the output of the StyleGAN2's generator that takes a latent code $w^\text{(i)}$ or $w^{+\text{(i})}$ and a stochastic noise input $n^\text{(i)}$. $\mathcal{L}_\text{loss}$ is a sum of all our loss functions:
\begin{align}
\label{eq:obj}
    \mathcal{L}_\text{loss} &=\sum_\text{j}^\text{[f, h, bg]}  \lambda_\text{p}^\text{(j)}\mathcal{L}_\text{per}^\text{(j)} \\
    &+ \lambda_\text{g} \mathcal{L}_\text{global} + \lambda_\text{i} \mathcal{L}_\text{ini} + \lambda_\varepsilon \mathcal{L}_\varepsilon + \lambda_s \mathcal{L}_\text{sim},
\end{align}
where $\lambda_{(\cdot)}$ are balancing weights, and $\text{[f, h, bg]}$ refer to face, hair, and background. We next explain each loss function.




\subsubsection{Loss Functions}
\label{sec:met_loss}
The objective of our $\W$ and $\WP$ optimization is to generate an output image that fits the corresponding $\iguide$ without appearing unrealistic. The main challenge is that our guide images are unrealistic and can contain various unwanted artifacts. We design our loss functions to be tolerant of the imprecise nature of our guide images.

\indent\textbf{Masked Perceptual Loss: }
This loss is based on Learned Perceptual Image Patch Similarity (LPIPS)~\cite{zhang2018perceptual}, which compares two input images in a \emph{deep feature space}. Similar to~\cite{saha2021LOHO,barbershop,kim2022styleyourhair}, we apply a binary mask to indicate regions of interest in the deep feature space of the two input images. We use $L_\text{LPIPS}(I_1, I_2; M)$ to denote this LPIPS computation between two images ($I_1$, $I_2$) with mask $M$. However, we observe that applying this masking operation after the feature computation is insufficient to disregard unwanted regions. This is due to the fact that LPIPS is a \emph{patch-based} similarity loss, and regions outside of the mask can still affect the loss computation. Based on this principle, we propose to apply additional \textbf{pre-masking} to mask out ambiguous regions in $\iguide$ that should not be trusted.
Specifically, our Masked Perceptual Loss is the following:
\begin{equation}
\label{eq:lper}
\resizebox{.9\hsize}{!}{
$\mathcal{L}_\text{per}^\text{i,roi} = \Lambda^\text{(i)}_\text{roi} L_\text{LPIPS}(O^\text{(i)} \odot \neg M^\text{(i)}_\text{roni}, \iguide^\text{(i)} \odot \neg M^\text{(i)}_\text{roni}; M^\text{(i)}_\text{roi}), $
}
\end{equation}
where $M_\text{roi}$ is the mask for region of interest (ROI), and $M_\text{roni}$ is the mask for regions of \emph{not} interest (RONI). $( . ) \odot \neg M_\text{roni}$ is a simple element-wise multiplication, which effectively excludes these regions from the loss by setting them to 0 in both input images. And $\Lambda^\text{(i)}_\text{roi}$ is a balancing weight in (i) viewpoint, which is set higher when the region of interest covers the original pixels of
$I_f$ (or $I_h$). 
For instance, the loss that attends to face in $\iguidef$ has a higher weight than the face loss in $\iguideh$.
We consider 3 regions of interest: $L_\text{per}^\text{f}$, $L_\text{per}^\text{h}$, and $L_\text{per}^\text{bg}$ for face, hair, and background regions. We refer to Section~\ref{sec:met_hall} and~\ref{sec:met_details} for details on these mask generation for $\W$ and $\WP$ optimization, respectively.

\indent\textbf{Global loss:}
This loss function attempts to match the overall appearance of the output to $\iguide$:
\begin{align}
\mathcal{L}_\text{global} &= L_\text{MSE}^{32}(O, \iguide).
\end{align}
$L_\text{MSE}^{32}$ is the mean square error computed on 32x32 downsampled input images. The goal of the downsampling is to reduce the effect of matching visible seams from higher resolution $\iguide$ because those seams become imperceptible once downsampled. This loss is inspired by a similar idea used for image super-resolution by Menon et al. \cite{pulse}.


\indent\textbf{Initial Value Loss:}
To prevent the output from deviating too far from the learned distribution and becoming unrealistic, we force the optimized latent code(s) $w$ and $w^+$ to be close to its initial value $w_0$ through L2 loss, as used in\cite{image2stylegan++}:
\begin{equation}
\begin{aligned}
\mathcal{L}_\text{ini} = \|w_{1:18} -w_0\|^2_2.
\end{aligned}
\end{equation}
We use the estimated mean of $\W$ latent space as initial values for optimizing $w$. This is computed by averaging many latent codes $w_i=f(z_i)$ drawn randomly through $z \sim \mathcal{N}(\mathbf{0}, \mathbf{I})$ and the StyleGAN's mapping network $f$.
Then the optimized latent code $w$ from the first stage becomes the initial value for the optimization in $\WP$ extended latent space.


\indent\textbf{Noise Regularization Loss:}
We also use the noise regularization loss ($\mathcal{L}_\varepsilon$) proposed in StyleGAN2~\cite{stylegan2} to ensure that the noise maps will be used for capturing the stochastic variations, and not for encoding the content meant to be captured by the latent code. We refer to StyleGAN2 for the details of this loss.

\indent\textbf{Latent Similarity Loss:}
To enable information sharing between the two guide images ($\iguidef$ and $\iguideh$) from two viewpoints, we force both their latent codes to be close together by using the following L2 loss, similarly to~\cite{chandran2021rendering}. 
\begin{equation}
\begin{aligned}
\mathcal{L}_\text{sim} = \|w^{(\text{face})} - w^{(\text{hair})}\|^2_2,
\end{aligned}
\end{equation}
where $w$ can be in $\W$ space during $\W$ optimization or replaced with $w^+$ during $\WP$ optimization. 

\subsubsection{Sharing Latent Code}
\label{sec:met_latshare}
In StyleGAN2, latent codes in early layers have been shown to correspond to high level concepts (\eg human pose) whereas the remaining layers capture low-level information (\eg colors)~\cite{stylegan, härkönen2020ganspace}. This motivates the optimization in $\WP$ extended space, as each layer can be optimized for different aspects of the image.
Similarly, we optimize latent codes to match guide images of the same person and hairstyle but in different poses.
Rather than optimizing two latent codes independently, we can enforce information sharing between the two poses by constraining the latent codes of the last few layers ($w_{l:18}$) to be the same. We set $l$ to be 4 in all our experiments, loosely based on the head rotation experiments shown in GANSpace~\cite{härkönen2020ganspace}.

\indent\textbf{Sharing $w$ Latent Code:}
Optimization in the space of StyleGAN2 latent code $\W$ is generally done as a single latent code ($	\mathbb{R}^{512}$), which are duplicated (18x) and fed into all the layers. In our case, we also optimize for a single latent code $w$ but only duplicate them for the first $l$ layers ($w_{1:l}$). The remaining layers ($w_{l:18}$) are interpolated from the current latent codes of each guide image with a coefficient $\alpha$.
\begin{align}
w_{l:18} &= \alpha w^\text{(face)}_{1:l} + (1-\alpha) w^\text{(hair)}_{1:l}.
\end{align}
We randomize this coefficient to make the latent codes stay within the space of $\W$. This random interpolation forces the latent code for each guide image to be partly similar while still producing realistic results from $\W$ latent space.

\indent\textbf{Sharing $w^+$ Latent Code:}
Optimization in $\WP$ extended space is straightforward, as latent code for each layer can already be separately optimized. We simply share the last few layers ($w_{l:18}$) of both $w^+$ latent codes during optimization. 
We refer to the Appendix~\ref{apx:optimization} for more details on the optimization process.

\subsection{\texorpdfstring{$\W$}{} Optimization: Hallucinate Missing Details}\label{sec:met_hall}
This stage aims to hallucinate details not currently visible in the guide images. The output images of this stage are $O_1^\text{face}$ and $O_1^\text{hair}$, which are essentially the projection of the unrealistic $\iguidef$ and $\iguideh$ into the real image distribution learned in StyleGAN2~\cite{stylegan2}. This is done by optimizing on $W$ latent space to fit each guide image with the objective function (Equation \ref{eq:obj}).

While our guide images ($\iguidef$ and $\iguideh$) capture the overall appearance of the desired outputs, they still lack details in certain regions. These unknown regions may correspond to unseen facial features that were occluded by the hair in $I_f$ or incomplete reference hair from $I_h$ caused by, \eg, a hat or image cropping. 
These details will need to be hallucinated and seamlessly blended with the rest of the image.

To accomplish this, we design our \textbf{pre-masking} in Masked Perceptual Loss ($\mathcal{L}_\text{per}^\text{f}$, $\mathcal{L}_\text{per}^\text{h}$, and $\mathcal{L}_\text{per}^\text{bg}$; Equation~\ref{eq:lper}) to include all the unknown regions that need to be hallucinated. Specifically, $M_\text{roni}^\text{f}$ represents the face region \emph{occluded} by the hair in $I_f$, \eg, the forehead behind the bangs or the ears that should become visible in the final output. Analogously, $M_\text{roni}^\text{h}$ represents the hair region occluded by other objects (\eg, a hat) or not visible in $I_h$ due to image cropping.
We set $M_\text{roi}^\text{f}$ and $M_\text{roi}^\text{h}$ to be segmentation masks for face and hair regions ($M_\text{f}$ and $M_\text{h}$), respectively. Both $M_\text{roi}^\text{bg}$ and $\neg M_\text{roni}^\text{bg}$ is set to be the background regions in $I_f$ not covered by the transferred hair ($M_\text{bg}$).
These masks are constructed by composing different semantic regions (union, intersection, etc.) from $I_f$ and $I_h$, detailed in Appendix \ref{sec:masks}
We note that $M_\text{roi}$ is applied on the deep feature maps, similar to~\cite{saha2021LOHO, barbershop, kim2022styleyourhair}, while $\neg M_\text{roni}$ is applied on raw RGB images. Using both masking techniques is crucial to our natural and seamless blending. The additional pre-masking allows the hair shape (or face shape) in the output to be different from $\iguide$'s and freely grow outward outside $M_\text{roi}$ if this leads to a more natural result (see Figure \ref{fig:ablation}).

\subsection{\texorpdfstring{$\WP$}{} Optimization: Recover Face-Hair Details}
\label{sec:met_details}
The output of the previous stage, $O_1^\text{face}$ and $O_1^\text{hair}$,  may still look different from the input person and not yet capture the hair details from the reference. This stage aims at refining these output images to better reproduce the hair details from $I_h$ and the rest from $I_f$. The optimization is done in the extended $\WP$ space with respect to $\{w^+, n\}$. In other words, we allow $w \in \mathbb{R}^{512}$ that was previously replicated to each layer to be optimized separately as $w^+ \in \mathbb{R}^{18 \times 512}$.
We denote $O_2^\text{face}$ and $O_2^\text{hair}$ to be the output of this stage generated by StyleGAN2 from our optimized latent codes in $\WP$ extended latent space.

We update the target images for the optimization from $\iguidef$ and $\iguideh$ to the more complete versions based on $O_1^\text{face}$ and $O_1^\text{hair}$ from the first stage. However, because the first stage aims at hallucinating new details into the pre-masking regions, it may lose the original texture details. To address this, we replace the known regions in $O_1^\text{face}$ and $O_1^\text{hair}$ with the correct details from the original $\iguide$ by setting
\begin{equation}\label{eq:updateI}
    I_\text{new\_guide} \leftarrow \iguide\odot M_\text{c} + O_1\odot \neg M_\text{c},
\end{equation}
where $M_c = M_\text{f} \cup M_\text{h} \cup M_\text{bg}$, which corresponds to the regions of interest in our Masked Perceptual Loss where we want to match with $\iguide$. In short, we simply copy the hallucinated parts from $O_1^\text{face}$ and $O_1^\text{hair}$, and combine them with the part in $\iguide$ that are known to be correct.

For our Masked Perceptual Loss, we use the same masking procedure as the previous stage (Section~\ref{sec:met_hall}). However, we observe that the projection from EG3D, which produces $\iguide$, generally is inferior to the known details from the original images. For example, $\iguideh$ contains more details on the hair than $\iguidef$ due to the latter being hallucinated from the 3D rotation by EG3D's rendering pipeline. To \emph{deprioritize} these inaccurate details, we blur both $\iguide$ and $O$ when computing the perceptual loss for the less accurate regions of interest. Concretely, we only blur images when computing $\mathcal{L}_\text{per}^\text{f}$ for $\iguideh$ optimization and when computing $\mathcal{L}_\text{per}^\text{h}$ for $\iguidef$ optimization.

\subsection{Pivotal Tuning Inversion}\label{sec:met_pti}
The purpose of this stage is to further refine the output images ($O_2^\text{face}$ and $O_2^\text{hair}$) by allowing the optimization of the StyleGAN's weights $\theta$ (while fixing the optimized latent code $w_\text{optimized}$). We closely follow the proposed optimization in PTI~\cite{pti} with the same objective function. However, we update the reconstruction loss to reflect the goal of our task, which is to match the original details from $I_f$ and $I_h$.
 \begin{align}
 \mathcal{L}_\text{pti} &= \sum_\text{i}^{[\text{face, hair}]}(L_\text{LPIPS}(O^\text{(i)}_\text{tune} \odot M^\text{(i)}_\text{raw}, I^\text{(i)}_\text{guide} \odot M^\text{(i)}_\text{raw}; M^\text{(i)}_\text{raw}) \nonumber \\
 &+ L_\text{MSE}^{32}(O^\text{(i)}_\text{tune}, I^\text{(i)}_\text{guide} ; \neg M^\text{(i)}_\text{raw})),
 \end{align}
where $O_\text{tune}$ is the generated image using the tuned weights $\theta$, the mask $M_\text{raw}^\text{face}$ is ($M_\text{bg} \cup M_\text{f}$), and $M_\text{raw}^\text{hair}$ is $M_\text{h}$.




\section{Experiments}
\label{sec:experiments}
In this section, we compare our method to state-of-the-art StyleYourHair~\cite{kim2022styleyourhair}, Barbershop~\cite{barbershop}, and LOHO~\cite{saha2021LOHO}.
Our evaluation criteria are i) user preference via a user study on a wide variety of scenarios, ii) hairstyle transfer quality, iii) hair reconstruction quality, and iv) how well the input face shape is preserved.
Section \ref{sec:ablation} presents ablation studies on our multi-view sharing latent optimization, optimization stage, and loss functions.


We use the official code of LOHO~\cite{saha2021LOHO} and Barbershop~\cite{barbershop} with the default configurations.
For \mbox{StyleYourHair}~\cite{kim2022styleyourhair}, we use the configuration where the hair reference is never flipped.

\label{sec:soa_comp}
\begin{figure}[t]
\begin{center}
\includegraphics[width=1\linewidth\vspace*{-0.6em}]{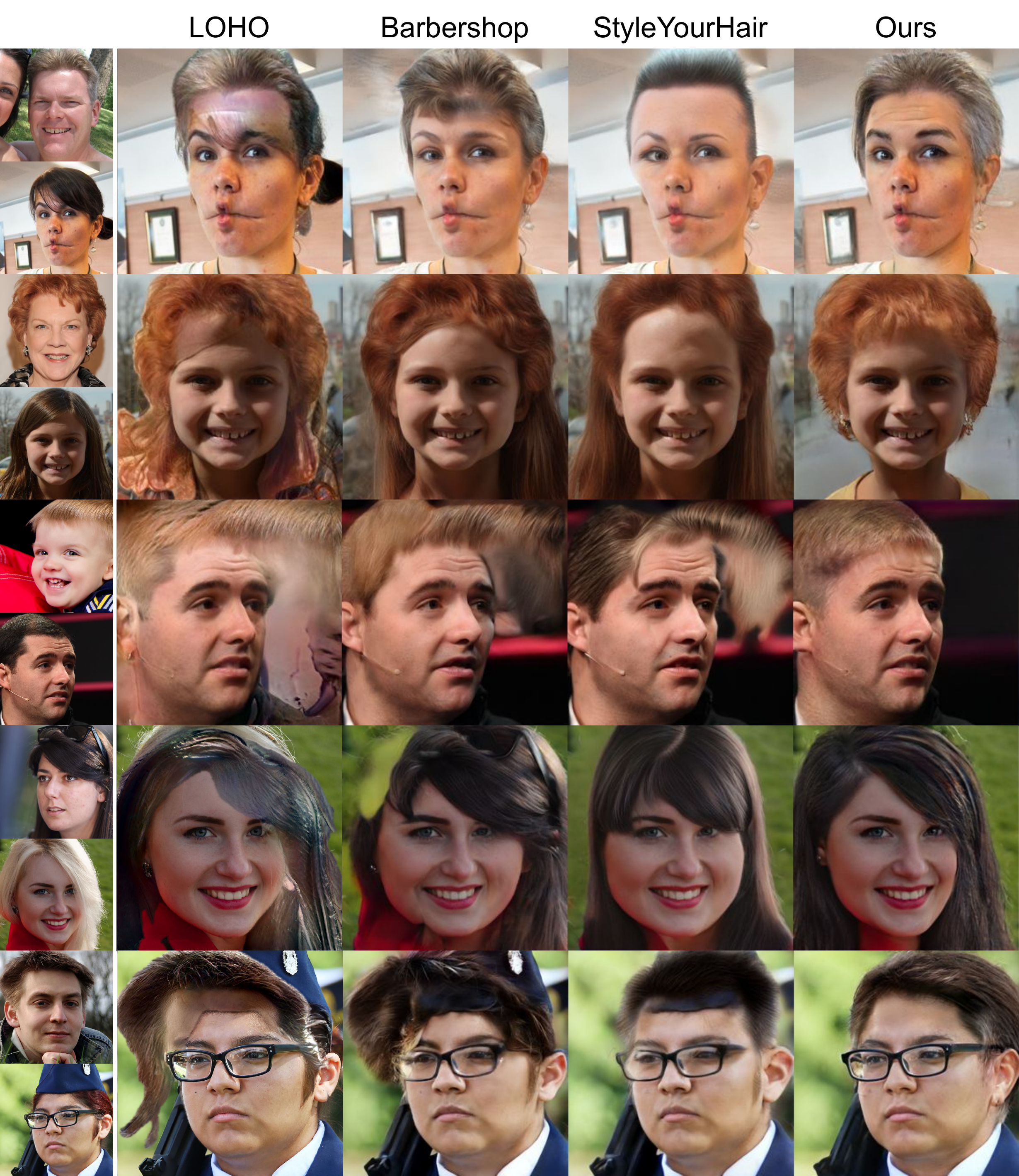}
\end{center}
  \caption{
  Comparison to current state-of-the-art methods for transferring hair from  1st column-top, to the face of 1st column-bottom.
  Our method can accurately transfer hairstyle even when the input face and hair are misaligned (3rd, 4th row). It can also hallucinate missing details such as a forehead that was previously occluded (1st row), or shorten the hair (2nd row). In general, LOHO and Barbershop struggle with pose misalignment, while StyleYourHair struggles to preserves the input's face shape and hair details.
  }
\label{fig:comparision1}
\vspace{-0.5cm}
\end{figure}


\begin{table*}[t]
\centering
\setlength{\tabcolsep}{3pt}
\resizebox{\textwidth}{!}{
\begin{tabular}{lcccccccccccccccc}
\toprule
 & \multicolumn{3}{c}{Test Datasets} & \multicolumn{13}{c}{FFHQ Scenario Breakdown } \\  \cmidrule(lr){2-4} \cmidrule(lr){5-17}
 & \multirow{5}{*}{\rotatebox[origin=c]{90}{\parbox[c]{2cm}{\centering All}}} & \multirow{5}{*}{\rotatebox[origin=c]{90}{\parbox[c]{2cm}{\centering FFHQ-P}}} & \multirow{5}{*}{\rotatebox[origin=c]{90}{\parbox[c]{2cm}{\centering FFHQ-S }}} & Easy & \multicolumn{4}{c}{Medium} & \multicolumn{7}{c}{Difficult} &
 \\ \cmidrule(lr){5-5} \cmidrule(lr){6-9} \cmidrule(lr){10-16}

&&&&\cellcolor[HTML]{EFEFEF}- &\cellcolor[HTML]{EFEFEF}\cmark &\cellcolor[HTML]{EFEFEF}- &\cellcolor[HTML]{EFEFEF}- &\cellcolor[HTML]{EFEFEF}- &\cellcolor[HTML]{EFEFEF}\cmark\cmark &\cellcolor[HTML]{EFEFEF}\cmark &\cellcolor[HTML]{EFEFEF}\cmark &\cellcolor[HTML]{EFEFEF}\cmark &\cellcolor[HTML]{EFEFEF}- &\cellcolor[HTML]{EFEFEF}- &\cellcolor[HTML]{EFEFEF}- &\multicolumn{1}{l}{\cellcolor[HTML]{EFEFEF}Pose Misalignment} \\
&&&&\cellcolor[HTML]{EFEFEF}- &\cellcolor[HTML]{EFEFEF}- &\cellcolor[HTML]{EFEFEF}\cmark &\cellcolor[HTML]{EFEFEF}- &\cellcolor[HTML]{EFEFEF}- &\cellcolor[HTML]{EFEFEF}- &\cellcolor[HTML]{EFEFEF}\cmark &\cellcolor[HTML]{EFEFEF}- &\cellcolor[HTML]{EFEFEF}- &\cellcolor[HTML]{EFEFEF}\cmark &\cellcolor[HTML]{EFEFEF}\cmark &\cellcolor[HTML]{EFEFEF}-&\multicolumn{1}{l}{\cellcolor[HTML]{EFEFEF}Needs Face Inpainting} \\
&&&&\cellcolor[HTML]{EFEFEF}- &\cellcolor[HTML]{EFEFEF}- &\cellcolor[HTML]{EFEFEF}- &\cellcolor[HTML]{EFEFEF}\cmark &\cellcolor[HTML]{EFEFEF}- &\cellcolor[HTML]{EFEFEF}- &\cellcolor[HTML]{EFEFEF}- &\cellcolor[HTML]{EFEFEF}\cmark &\cellcolor[HTML]{EFEFEF}- &\cellcolor[HTML]{EFEFEF}\cmark &\cellcolor[HTML]{EFEFEF}- &\cellcolor[HTML]{EFEFEF}\cmark &\multicolumn{1}{l}{\cellcolor[HTML]{EFEFEF}Needs BG Inpainting} \\
&&&&\cellcolor[HTML]{EFEFEF}- &\cellcolor[HTML]{EFEFEF}- &\cellcolor[HTML]{EFEFEF}- &\cellcolor[HTML]{EFEFEF}-&\cellcolor[HTML]{EFEFEF}\cmark &\cellcolor[HTML]{EFEFEF}- &\cellcolor[HTML]{EFEFEF}- &\cellcolor[HTML]{EFEFEF}- &\cellcolor[HTML]{EFEFEF}\cmark &\cellcolor[HTML]{EFEFEF}- &\cellcolor[HTML]{EFEFEF}\cmark &\cellcolor[HTML]{EFEFEF}\cmark &\multicolumn{1}{l}{\cellcolor[HTML]{EFEFEF}$I_h$ wears a hat} \\ \midrule

LOHO          & 8.8& 9.6& 8.4& 12.0& 5.3& 4.0& 13.3& 8.0& 5.3& 5.3& 4.0& 10.7& 5.3& 13.3& 14.7   &       \\
Barbershop    & 16.4& \underline{13.8}& 17.8& 17.3& \underline{21.3}& \underline{13.3}& 21.3& 18.7& \underline{17.3}& 17.3& 20.0& 12.0& 18.7& \underline{14.7}& \underline{21.3}      &   \\
StyleYourHair & \underline{18.5}& 12.4& \underline{21.6}& \underline{25.3}& \underline{21.3}& 10.7& \underline{22.7}& \underline{33.3}& 10.7& \underline{24.0}& \underline{28.0}& \underline{14.7}& \underline{32.0}& \underline{14.7}& \underline{21.3}   \\
Ours        & \textbf{56.2} &  \textbf{64.2}& \textbf{52.2}& \textbf{45.3}& \textbf{52.0}& \textbf{72.0}& \textbf{42.7}& \textbf{40.0}& \textbf{66.7}& \textbf{53.3}& \textbf{48.0}& \textbf{62.7}& \textbf{44.0}& \textbf{57.3}& \textbf{42.7}  & \\ \bottomrule

\end{tabular}}
\caption{\label{table:user_study} 
User study results on hairstyle transfer (percentage of user preferring each method). Our method outperforms state-of-the-art hairstyle transfer methods on FFHQ datasets in all challenging scenarios. A total of 450 pairs are used in this study, 150 pairs in FFHQ-P and 300 in FFHQ-S. For each pair, we asked 3 unique participants to select the best result for hairstyle transfer.}

\end{table*}

\subsection{Qualitative Comparison}
\label{sec:soa_comp_qual}
We provide a qualitative comparison to StyleYourHair~\cite{kim2022styleyourhair}, Barbershop~\cite{barbershop}, and LOHO~\cite{saha2021LOHO} in Figure \ref{fig:comparision1}.
We observe that LOHO and Barbershop often struggle to fit the reference hair accurately when the poses are not well aligned (rows 3-5), resulting in various artifacts, such as remnants of the original hair or wrong placement of the target hair. Barbershop performs well in preserving hair texture when the poses are similar. However, it falls short when handling challenging cases, such as transitioning from a long to a short hairstyle or removing bangs.

StyleYourHair can produce more realistic results in unaligned cases than LOHO and Barbershop. However, the hair details often look different from the reference hairstyle (row 5). We also observe that StyleYourHair may perform poorly when the pose difference becomes too large (row 3), as also shown in our user study for pose in Table \ref{table:user_study}.
In contrast, our method can transfer hairstyles convincingly, regardless of the misalignment, while still preserving the original face shape and hair details. We refer to Appendix~\ref{apx:other_results} for more results.


\subsection{User Study}
\label{sec:user_study}
Qualitative results can be misleading and biased, so we conducted a user study using Amazon Mechanical Turk on hairstyle transfer results using randomly selected pairs grouped into various difficulty levels. We compare results of our method with the current state of the art: StyleYourHair, Barbershop, and LOHO.
Each participant was shown an input face, marked as `Face', and a reference hair, marked as `Hair', and asked to pick only one output that best accomplishes the task of transferring the hairstyle from image `Hair' to the person in image `Face'. The output row consists of four images from each method \emph{in random order}. Each task was evaluated by 3 different participants.
All images were in 256x256 resolution.

\subsubsection{Datasets}
For the user study, we construct two challenging benchmarks: FFHQ-P and FFHQ-S, from the test set of Flickr-Faces-HQ dataset (FFHQ)~\cite{stylegan}.

\textbf{FFHQ-P} contains 150 random input face-hair pairs from FFHQ, covering yaw differences in ranges of $[0-15), [15-30), \ldots , [75-90)$, with 25 pairs in each range.


\textbf{FFHQ-S} contains 300 pairs, categorized into 12 different configs with varying levels of difficulty (Table~\ref{table:user_study}). Each config contains 25 random input pairs and is a combination of four possible scenarios (see details in Appendix~\ref{apx:create_dataset}):
\begin{compactitem}
    \item Pose Misalignment: When the yaw difference is between [15-30) or [30-45), a single checkmark or double checkmarks are used, respectively in Table~\ref{table:user_study}.
    
    \item Needs Face Inpainting: This includes scenarios that require hallucinating parts of the original face, \eg, inpainting the forehead to remove bangs.
    This is challenging because the identity can easily change from the hallucination. We detect such scenarios based on the face/hair regions in $I_f$ and $I_h$ (Appendix \ref{apx:create_dataset}).
    
    \item Needs BG Inpainting: This includes scenarios where the hair shape becomes smaller and requires background inpainting. We detect such scenarios automatically based on the hair regions in $I_f$ and $I_h$ (Appendix \ref{apx:create_dataset}).
    
    \item $I_h$ Contains Hat: This represents scenarios where the hair reference is not fully visible in $I_h$. We detect such scenarios based on the hat region in $I_h$.
 \end{compactitem}


\subsubsection{Results}
Using all test pairs, the participants preferred our results 56.2\% of the time, whereas StyleYourHair’s, Barbershop’s, and LOHO’s results were selected for 18.5\%, 16.4\% and 8.8\%, respectively. Our method perform the best in both FFHQ-P (64.2\%) and FFHQ-S (52.2\%), and in all configurations in Table~\ref{table:user_study}-\ref{table:user_pose}. 
%
%

\begin{table}[t]
\begin{center}
\setlength{\tabcolsep}{0.3em}
\resizebox{\columnwidth}{!}{
\begin{tabular}{lcccccc}
\toprule
           & \multicolumn{6}{c}{Pose Difference Range (FFHQ-P)}  \\ 
           & [0,15)  & [15,30) & [30,45) & [45,60) & [60,75) & [75,90) \\ \midrule
LOHO       &10.7 & 5.3  & 10.7 &12.0  & 8.0 &  10.7              \\
Barbershop &\underline{24.0}& 13.3  &\underline{12.0}  & \underline{8.0} & 10.7 & \underline{14.7}              \\
StyleYourHair  &17.3&  \underline{16.0}&  \underline{12.0}&  6.7& \underline{14.7}&  8.0     \\
Ours    & \textbf{48.0} & \textbf{65.3} & \textbf{65.3} & \textbf{73.3} & \textbf{66.7} & \textbf{66.7}    \\ \bottomrule
\end{tabular}}
\caption{ \label{table:user_pose} User study on pose-invariant hairstyle transfer. Our method outperforms others on all pose difference ranges.
}
\end{center}
\vspace{-0.5cm}
\end{table}

\begin{table}[t]
\begin{center}
\setlength{\tabcolsep}{0.5em}
\resizebox{\columnwidth}{!}{
\begin{tabular}{lcccc|c}
\toprule
          & \multicolumn{4}{c|}{Hair Reconstruction Metrics} & Face Shape   \\ 
             & PSNR $\uparrow$  & SSIM $\uparrow$ & LPIPS $\downarrow$ & FID $\downarrow$ & RMSE $\downarrow$ \\ \midrule
LOHO         &25.76  & 0.86 & \underline{0.07} & \textbf{10.40} &  15.84            \\
Barbershop  & \textbf{29.18}  & \textbf{0.89} &\textbf{0.05} & \underline{10.46} &  17.25            \\
StyleYourHair  & 26.89 & 0.87& 0.09 & 10.93  &  \underline{14.37}            \\
Ours     & \underline{27.84} & \underline{0.88} & \underline{0.07} & 10.85 & \textbf{10.89}   \\ \bottomrule
\end{tabular}}
\caption{ \label{table:both_recon} Hair reconstruction results in the self-transfer experiment (Section \ref{sec:hairrecon}), and RMSEs between facial landmarks detected on the input and output images.
(Section \ref{sec:soa_comp_face}).
}

\end{center}
\vspace{-0.5cm}
\end{table}


\subsection{Quality of Hair Reconstruction}
\label{sec:hairrecon}
Following MichiGAN and LOHO, we perform a self-transfer experiment where we set both the input face and input reference hair to be the same and evaluate the reconstruction accuracy 
using various metrics: PSNR, SSIM, IPIPS, FID~\cite{fidheusel2017gans}. 
For every method, the hair region of the output will be blended back to the input image to ensure that the difference in score only comes from the hair.

The scores of our method, LOHO, Barbershop, and StyleYourHair are reported in Table~\ref{table:both_recon}.
Unsurprisingly, when there is no misalignment between input face and hair, Barbershop generally achieves excellent hair reconstruction quality. Like ours, StyleYourHair is specifically designed to be pose-invariant, but their hair reconstruction quality seems to suffer greatly, as shown in our comparison (row 5 of Figure~\ref{fig:comparision1}). Our method achieves better performance at hair reconstruction than StyleYourHair, but also suffers from similar drawbacks.

\begin{figure*}[t]
\centering
\includegraphics[width=1\linewidth\vspace*{-0.4em}]{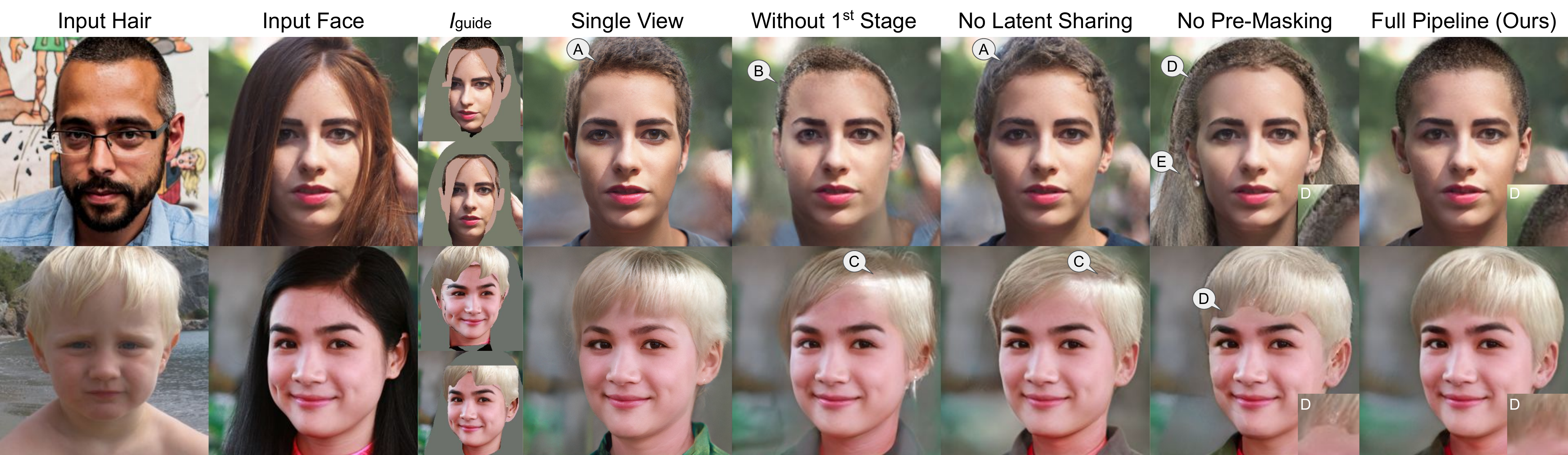}
  \caption{Ablation study of optimization stages and pre-masking. These ablated versions of our pipeline may produce results that (A) look realistic, but
fail to match the $I_h$’s hairstyle; (B, C) have unlikely hair shapes or structures; (D) have sharp boundaries; and (E) contain an incomplete background.
  }
\label{fig:ablation}
\vspace{-0.5cm}
\end{figure*}

\subsection{How Well Is Face Shape Preserved?}
\label{sec:soa_comp_face}
We also propose a novel evaluation metric for hairstyle transfer that focuses on the ability to preserve facial shape of the original person. We accomplish this by comparing detected keypoints of the input face $I_f$ and those on the output using a simple Root-Mean-Square Error (RMSE). These keypoints were detected using an off-the-shelf library Dlib~\cite{dlib09}, and we only used keypoints on the facial contour (ID 0-16) for this evaluation.

For this evaluation, we randomize additional 1,550 FFHQ test pairs into our FFHQ-P and FFHQ-S datasets (2,000 images in total). 
Table~\ref{table:both_recon} shows that our method outperforms others with an RMSE of 10.89, which is 24.2\% lower than StyleYourHair's (the second best), 36.9\% lower than Barbershop's and 31.3\% lower than LOHO's.

\subsection{Ablation Studies} \label{sec:ablation}
Here we assess the importance of each component in our pipeline. The results are shown in Figure~\ref{fig:ablation}. We test 4 ablation configurations: 
\textbf{i)} by using only $\iguidef$ from a single view, 
\textbf{ii)} without using the hallucination stage (Section \ref{sec:met_hall}), 
\textbf{iii)} without using the latent sharing structure (Section \ref{sec:met_latshare}), and 
\textbf{iv)} without using pre-masking in $\mathcal{L}_\text{per}$ (Section \ref{sec:met_loss}).

Optimizing with a single view guide $\iguidef$ (Config i) yields inaccurate hairstyles (Figure~\ref{fig:ablation}-A). 
Without optimization in W (Config ii), the method produces unrealistic hair results with various artifacts, \eg, sharp edges, unnatural hair shapes (Figure~\ref{fig:ablation}-B) or structures (Figure~\ref{fig:ablation}-C). This is because the second stage, which has higher image fitting capability, tries to fit the initial rough estimation $\iguide$. 
Without the latent sharing structure, the hair detail cannot be shared from the reference hair view ($\mathcal{L}_\text{sim}$ alone is not strong enough to force consistency) (Figure~\ref{fig:ablation}-A), resulting in inaccurate hair colors or structures (Figure~\ref{fig:ablation}-C). 


Without pre-masking in $\mathcal{L}_\text{per}$, the boundary of the face and hair regions is forced to be the same as in $\iguide$, leading to visible and sharp seams between the face and hair or between different facial features. Here the optimizer fails to refine the face-hair boundary to make the results look natural (Figure~\ref{fig:ablation}-D), and the original details, such as the background from $I_f$ cannot be seamlessly blended (Figure~\ref{fig:ablation}-E).

\begin{figure}
\centering
\includegraphics[width=1\linewidth\vspace*{-0.2em}]{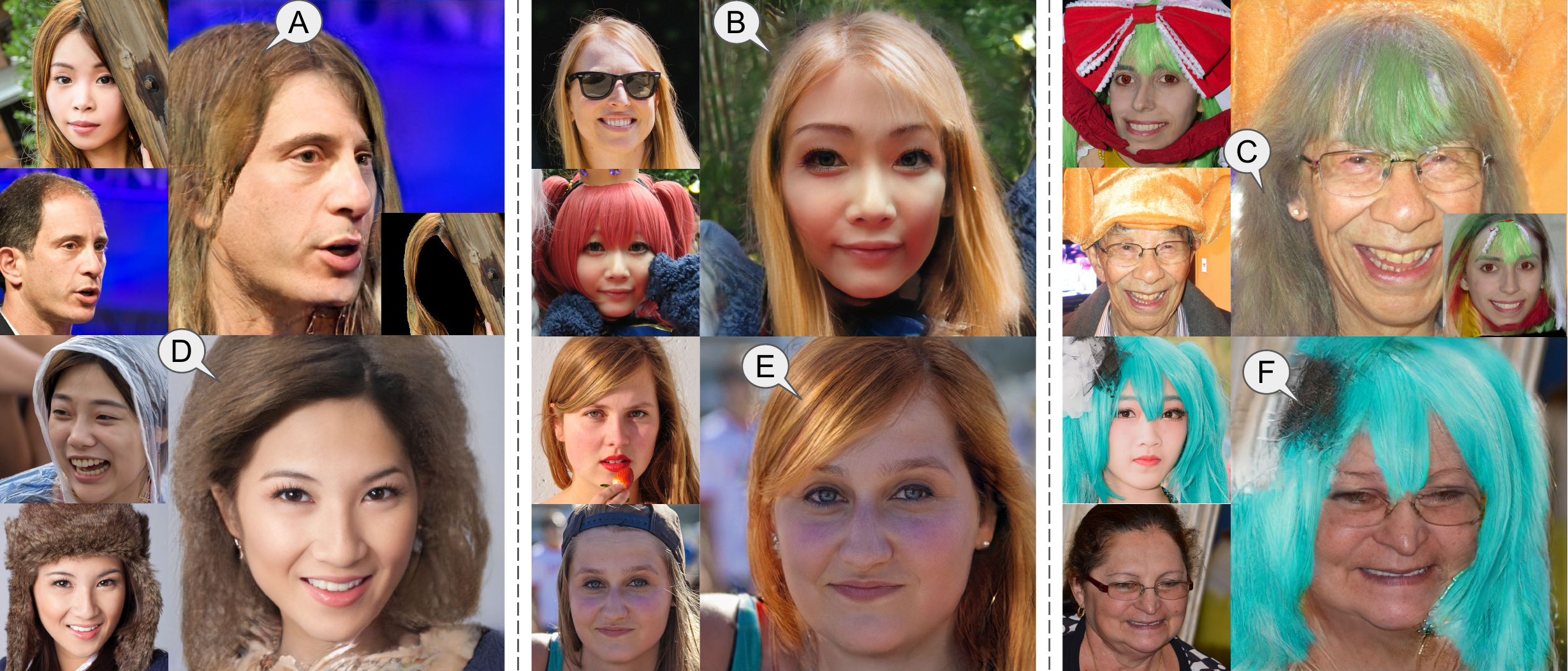}
  \caption{Failure cases; (A) the semantic regions are incorrect; (B) the facial keypoint is incorrect; (C) the EG3D projection is incorrect; (D) the reference hair color is too similar to the background color; (E) lighting looks unnatural; (F) eccentric hairstyle.
  }
\label{fig:fail}
\end{figure}

\section{Conclusion and Limitations}
\label{sec:conclusion}
We have presented a flexible and effective hairstyle transfer system that can handle
a variety of challenging in-the-wild scenarios and produce perceptually convincing hairstyle transfer results.
Our user study shows that human evaluators prefer our results over previous methods across all tested scenarios.
Nevertheless, our method can still fail in certain scenarios, for example, when the face and hair are too eccentric (Figure \ref{fig:fail}). We refer to Appendix~\ref{apx:fail} for more details.

{\small
\bibliographystyle{ieee_fullname}
\bibliography{11_references}
}
\clearpage
\appendix

\section*{Appendix}
In this appendix, we provide additional explanations, experiments, and results:
\begin{compactitem}
    \item Section~\ref{sec:det}: Implementation details of our guide image, mask construction, and optimization.
    \item Section~\ref{apx:create_dataset}: Details of our FFHQ-S test set construction.
    \item Section~\ref{apx:other_exp}: Additional experiments.
    \item Section~\ref{apx:other_abla}: Additional ablation studies.
    \item Section~\ref{apx:hairnet}: Comparison to a concurrent work.
    \item Section~\ref{apx:other_results}: Additional real-image hairstyle transfer results and examples from our user study.
    \item Section~\ref{apx:fail}: Failure cases.
    \item Section~\ref{apx:negative}: Negative societal impact
     
\end{compactitem}

\section{Implementation Details} \label{sec:det}
For guide image construction and image blending, we rely on 2D facial keypoints predicted by Dlib library~\cite{dlib09} and semantic regions predicted from a pretrained segmentation network~\cite{facegithub, yu2018bisenet}. The semantic output contains 19 classes, but we group them into 6 classes: face, ear, nose, neck, hair, and background.
After both input face $I_f$ and reference hair $I_h$ are aligned in both viewpoints, we denote by $H_\text{hair}, H_\text{face}, ...$ the semantic regions of $I_h$ for the hair, face, or other parts, and analogously by $F_\text{hair}, F_\text{face}, ...$ for the parts in $I_f$.

\subsection{Face-Hair Alignment}\label{sec:det_pre}
The purpose of this step is to create pose-aligned versions of  $I_f$ and $I_h$. That is, a new $I_f$ in the head pose of $I_h$ and a new $I_h$ in the head pose of $I_f$. These pose-aligned $I_f, I_h$ will be used to construct guide images.
We utilize EG3D~\cite{eg3d} for this task and use uniform scaling and translation to match their faces' widths and positions. We first explain the process to warp $I_h$ to match the head pose of $I_f$. The other direction from $I_f$ to $I_h$ will be done similarly with a small change, discussed afterward.
\subsubsection{EG3D Warping}
\begin{figure}[t]
\centering
\includegraphics[width=1\linewidth]{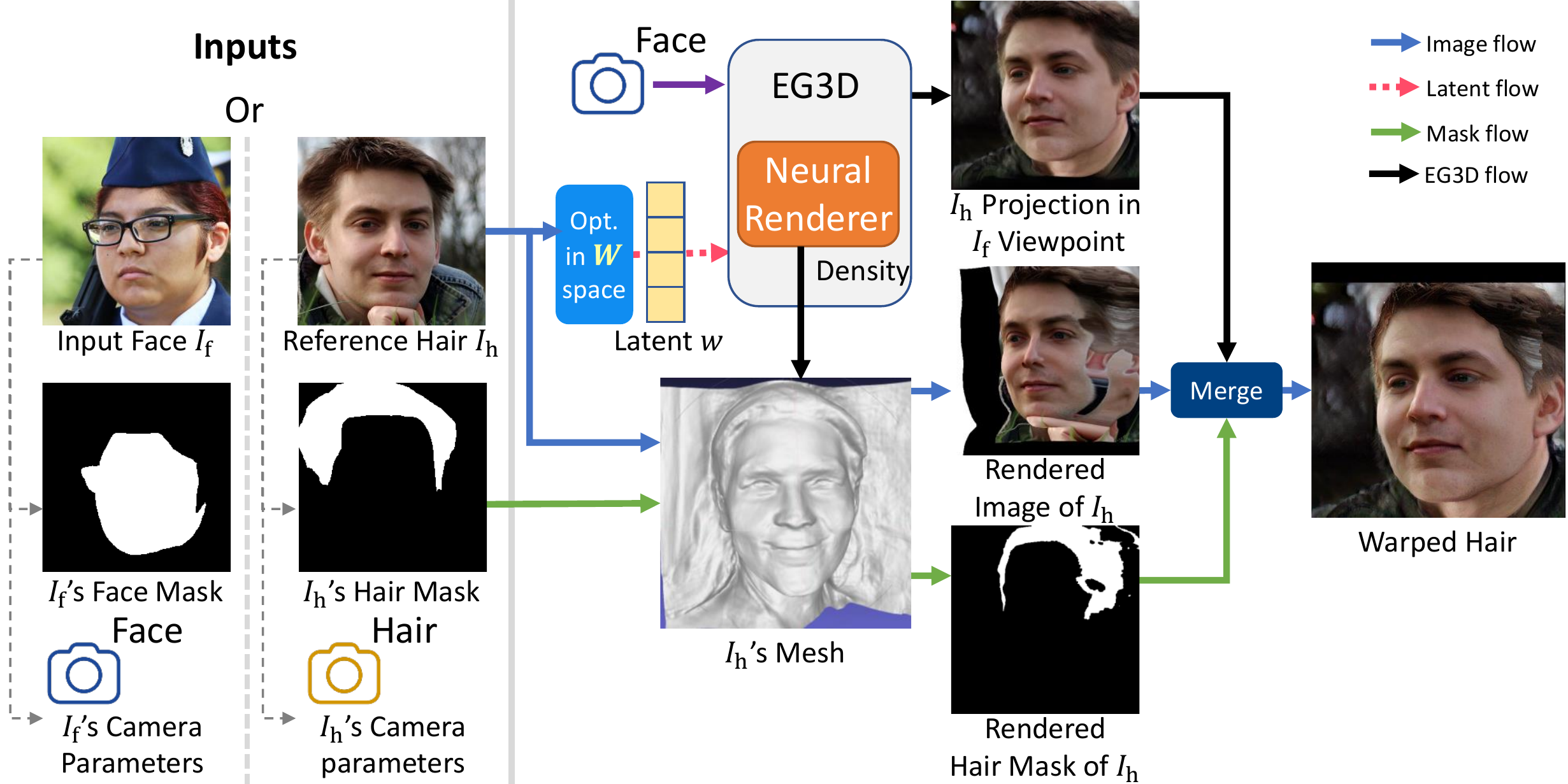}
  \caption{\textbf{Overview of EG3D warping:}
  We rotate $I_h$ to the pose of  $I_f$, and vice versa (not visualized in the diagram). This is done by combining the projection and geometry consistency of EG3D with the texture detail of the input image.}
  
\label{fig:eg3dwarp-pipeline}
\end{figure}

We use EG3D~\cite{eg3d} to rotate $I_h$ to match $I_f$'s pose. While EG3D projection can provide consistent geometry, the original details are often not well preserved. To fix this, we present a warping method that directly uses the original hair pixels by utilizing the EG3D estimated geometry, as shown in Figure \ref{fig:eg3dwarp-pipeline}. We preprocess an input image and determine the camera pose using the EG3D-proposed technique. We use the official code of EG3D with their ffhq512-64.pkl checkpoint.

\indent\textbf{Mesh Retrieval: }
We project $I_h$ into EG3D's $\W$ latent space using PTI~\cite{pti}, ignoring the hat region, which can lead to inaccurate segmentation after projection. Then, we use the Marching cube algorithm~\cite{lorensen1987marching} to construct a triangle mesh from the volume density of EG3D. We assign colors to the mesh by reprojecting the mesh onto $I_h$ and use the pixel colors from $I_h$.
We store the triangles that belong to the hair region of $I_h$ in a set $\Psi$. This will be used to determine which pixels in the mesh, after being warped into the target viewpoint, correspond to the original hair pixels from $I_h$.

\indent\textbf{Target-Viewpoint Rendering:}
We render the mesh in the 512x512 resolution in $I_f$'s pose and replace all pixels outside the hair region of $I_h$ (not in $\Psi$) with the $I_h$'s projection that is warped to the $I_f$'s pose by EG3D.  
We repeat the same process to warp $I_f$ to $I_h$, but switching their roles and change the set $\Psi$ to contain the face region of $I_f$ instead.

\subsubsection{Uniform Scaling and Translation:} 

We align $I_h$ with $I_f$ by uniform scaling and translation to match the faces' widths and centers.
The face width is calculated as the difference in the x-coordinate between the  right-most ($k_{16}$) and left-most ($k_{0}$) keypoints. The x and y center coordinates are computed separately. The x-center is the x coordinate of $(k_0 + k_{16})/2$ and the y-center is the y coordinate of $((k_0 + k_{16})/2 + k_8)/2$. This improves alignment of faces with larger pose differences.

\subsection{Color Fill-In in the Guide Image Construction} \label{sec:guide}
\begin{figure}[t]
\begin{center}
\includegraphics[width=1\linewidth\vspace*{-0.5em}]{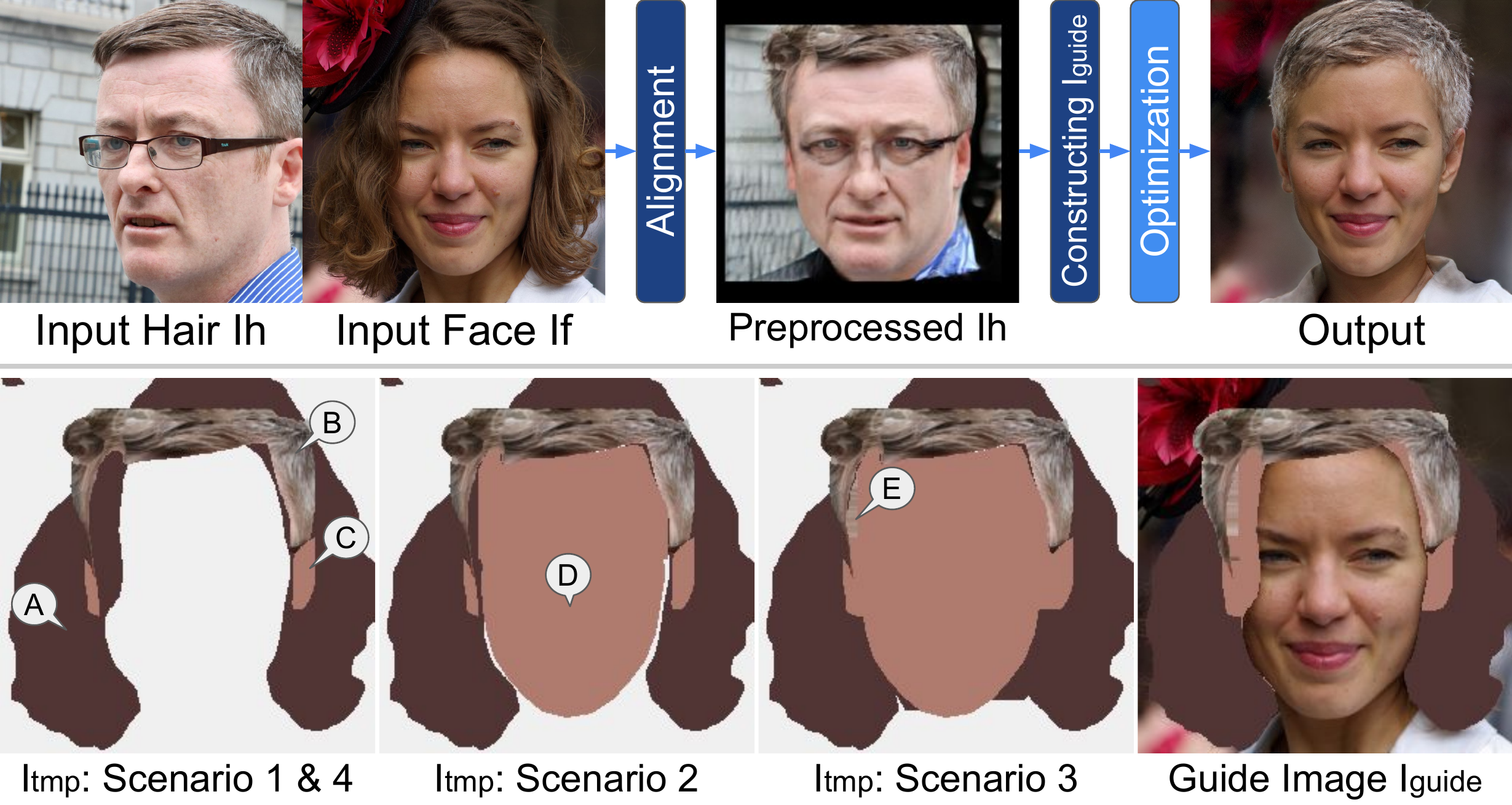}
\end{center}
  \caption{
  \textbf{Guide image construction.} We create a guide images $\iguide$ with a simple cut-and-paste of the hair from $I_h$ onto $I_f$ with a few modifications to handle the four scenarios in Section~\ref{sec:guide}. (We also create another guide image by transferring the face from $I_f$ to $I_h$, but not visualized by this diagram.) We solve this by first creating a temporary canvas image $I_\text{tmp}$ to remove the original hair in $I_f$ (Scenario 1; A,B), transfer the visible ears from $I_h$ to $\iguide$ (Scenario 4; C), and remove $I_f$'s bangs (Scenario 2; D) while preserving the face width of $I_f$ (Scenario 3; E). Then, we combine $I_\text{tmp}$ with $I_f$ and $I_h$ to create $\iguide$. }
\label{fig:guide}
\end{figure}

To construct $\iguide$, there are some corner cases that need to be properly handled, such as when the existing hair in the input is larger than the reference hair. Fortunately, addressing most corner cases amounts to handling the following four scenarios, as shown in Figure~\ref{fig:guide}.

\begin{enumerate}
    \item If the existing hair shape in $I_f$ is larger than the reference hair in $I_h$, fill the hair region not overlapped by the reference hair with the average background color.
    \item If $I_f$ has bangs on the forehead, remove them by filling that region with the average skin color.
    \item If $I_f$ has a narrower face than $I_h$, the guide image should retain the face width of $I_f$ and fill in both sides next to the face with the hair color of $I_h$.
    \item If the ears are visible in $I_h$, transfer the ear regions to the guide image and fill those regions with the skin color of $I_f$.
\end{enumerate}
We first create a temporary canvas image $I_\text{tmp}$ as follows.
We fill the region $F_\text{hair}$ in $I_\text{tmp}$ with the average color of the pixels within $F_\text{bg}$ (background) in $I_f$ (Scenario 1), and copy $H_\text{hair}$ in $I_h$ to $I_\text{tmp}$.
Then, if $H_\text{ear}$ exists (Scenario 4), fill the region $H_\text{ear}$ in $I_\text{tmp}$ with the skin color of $I_f$.
We approximate the skin color by averaging the nose pixels of the $I_f$ in $F_\text{nose}$.
Then, to remove the existing hair (Scenario 2), we fill the face area of $I_\text{tmp}$ defined by the area above keypoints $k_0-k_{16}$ with the $I_f$'s skin color. This face area is denoted by $F_\text{face}^\text{k}$.
We then fill $I_\text{tmp}$ in the region $F_\text{face}^\text{k} \cap (H_\text{face} \cup H_\text{neck})$ with the $I_f$'s skin color.
In the case where $I_f$ has a narrower face (Scenario 3), we fill $(H_\text{face} \cup H_\text{neck}) - F_\text{face}^\text{k}$ with the hair color of $I_h$ in a row-by-row basis.
After $I_\text{tmp}$ has been created, our $I_\text{guide}$ is constructed by first setting $I_\text{guide} = I_f$, then copying the content in the region $F_\text{hair}$ of $I_\text{tmp}$ to $I_\text{guide}$. Lastly, when some part of $H_\text{hair}$ overlaps with $F_\text{face}$, it is unclear whether the overlapped region should be hair or face in $I_\text{guide}$. To solve this, we update $H_\text{hair}$ by removing any region of $H_\text{hair}$ that lies outside of the face region defined by the detected keypoints. With this updated $H_\text{hair}$, we copy the hair of $I_h$ in this region to $I_\text{guide}$ to finish its construction.

\subsection{Mask Construction} \label{sec:masks}
\begin{figure}[t]
\begin{center}
\includegraphics[width=1\linewidth\vspace*{-0.5em}]{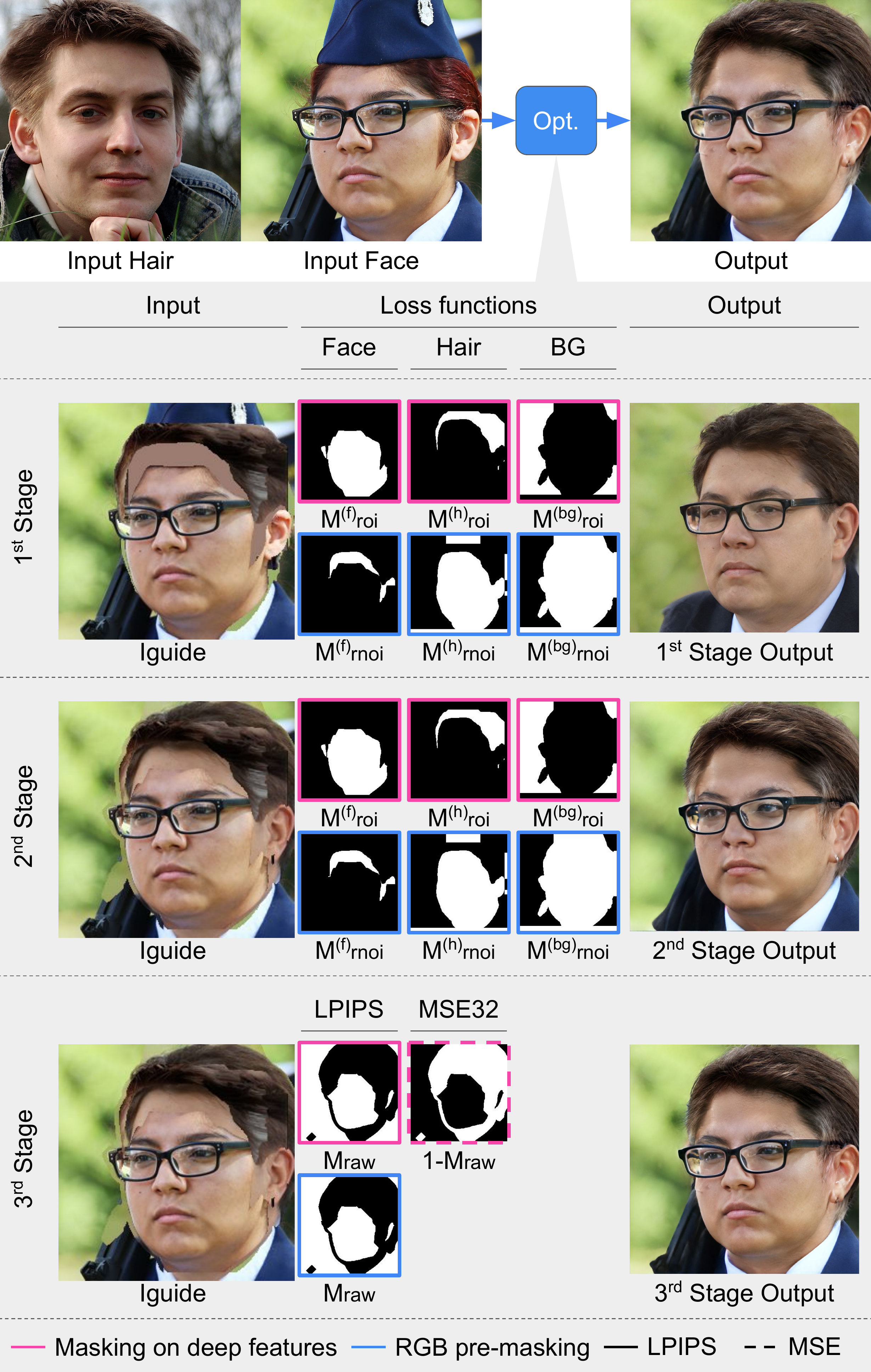}
\end{center}
  \caption{Masks used in each loss function. Masks used on the deep features in LPIPS are in pink frames, and masks used for RGB-premasking are in blue frames.
    }
\label{fig:masks}
\end{figure}

We show the masks used in each loss function in Figure~\ref{fig:masks}.

$\bm{M_\text{roi}^\text{f}; M_\text{f}}$ represents the face region of $I_\text{guide}$, computed by $F_\text{face} -  H_\text{hair} -  H_\text{hat}$. Additionally, we erode the region in $M_\text{f}$ that is higher than the eyebrows (5 pixels above the highest keypoints) using 5 iterations.

$\bm{M_\text{roi}^\text{h}; M_\text{h}}$ represents the hair region of $I_\text{guide}$, computed by $\text{\emph{erode}}(H_\text{hair}, 5)$. (I.e., eroding $H_\text{hair}$ using 5 iterations)

$\bm{M_\text{roi}^\text{bg}; M_\text{bg}}$ represents the background region in $I_f$ that is not covered by the transferred hair, computed by $F_\text{bg} - \text{\emph{dilate}}(H_\text{hair} \cup M_\text{out}, 5)$,
where $M_\text{out}$ represents out-of-the-frame regions. (Suppose, for example, $H_\text{hair}$ extends down to the bottom edge of $I_h$, the entire region below it in $ M_\text{out}$ will be marked 1).
Because there is no accurate background in the $I_h$ viewpoint, every pixel in $M^\text{hair}_\text{bg}$ is zero.

$\bm{M_\text{rnoi}^\text{f}}$ represents the face region in $I_f$ that was previously occluded  but should be visible in the final output, computed by $F_\text{face}^\text{k} + H_\text{ear} - M_\text{f}$.

$\bm{M_\text{rnoi}^\text{h}}$ represents the hair region that was previously occluded by other objects or not visible due to image cropping, computed by $H_\text{hat} \cup H_\text{face} \cup H_\text{neck} \cup M_\text{out}$.

$\bm{M_\text{c}}$ represents the regions in $\iguide$ that were copy-pasted from $I_f$ or $I_h$ (including warped pixels), computed by $\text{\emph{erode}}(M_\text{f} \cup M_\text{h} \cup (F_\text{bg} \cap O^1_\text{bg}), 5)$, where $O^1_\text{bg}$ represents the background region in $O_1$.

$\bm{M_\text{raw}}$ represents the regions in $\iguide$ that were copy-pasted from $I_f$ or $I_h$ with the original pixel content. 
The mask $M_\text{raw}^\text{face}$ is computed by $\text{\emph{erode}}(M_\text{bg} \cup M_\text{f}, 10)$, 
and $M_\text{raw}^\text{hair}$ is $\text{\emph{erode}}(M_\text{h}, 5)$.

\subsection{Regularization Losses}

This section elaborates on the regularization losses described in prior work.

\indent\textbf{Noise Regularization Loss:}
This loss proposed in StyleGAN2~\cite{stylegan2} is used ensure that the noise maps capture only the stochastic variations, by encouraging the optimized noise maps to be normally distributed via minimizing the normalized spatial autocorrelation:
\begin{equation}
\resizebox{.9\hsize}{!}{$\mathcal{L}_\varepsilon = \sum_{i, j} \left(\text{mean}(n_i^j \odot H(n_i^j))^2 + \text{mean}(n_i^j \odot V(n_i^j))^2\right)$,}
\end{equation}
where $H(\cdot)$, $V(\cdot)$ shift the noise map horizontally/vertically by one pixel with wrap-around edges. And for each noise map $n_i$, the autocorrelation is computed for different downscaled versions $n_i^0, n_i^1,...$ down to the 8x8 resolution.

\indent\textbf{PTI Regularization Loss:} 
This loss proposed in PTI~\cite{pti} restricts any change in the latent space to a small area. In each iteration, we sample $w_z$ from $\W$ space, then create an interpolated latent code $w_r$ between $w_z$ and the optimized latent code $w$ with an $\alpha$ parameter.

\begin{align}
w_r = w_\text{optimized} + \alpha\frac{w_z - w_\text{optimized}}{\left \| w_z - w_\text{optimized} \right \|_2}
\end{align}

To compute the final loss value, we feed the latent code $w_r$ into 
the original StyleGAN to produce $R$ and into a weight-tuned StyleGAN to produce $R^{*}$.
\begin{align}
\mathcal{L}_\text{r} &= L_\text{LPIPS}(R , R^{*}) + L_\text{MSE}(R, R^{*})
\end{align}

\subsection{Optimization Details} \label{apx:optimization}
For our $\W$ and $\WP$ latent optimization, we use Adam optimizer~\cite{adam} with $(\beta_1, \beta_2)=(0.9, 0.999)$. 
We run the optimization for 1,000 iterations in the first stage, and 500 iterations in the second stage with the same learning schedule used in StyleGAN2~\cite{stylegan2}. The learning rate is ramped up linearly from 0 to 0.1 during the first 5 percent of iterations (50/25) and ramped down with a cosine schedule during the last 25 percent of iterations (250/125). 
The initial $w_0$ used in the first stage is computed by averaging 10,000 latent codes (Section \ref{sec:met_loss}).

The parameter $\Lambda^\text{(i)}$, which is used to scale $\mathcal{L}_\text{per}^\text{[f/bg]}$ in $I_f$ viewpoint and $\mathcal{L}_\text{per}^\text{h}$ in $I_h$ viewpoint, is set to 6 in the first stage, 4 in the second stage. (The other losses are scaled by 1).
The parameters ($\lambda_\text{p}^\text{f}, \lambda_\text{p}^\text{h}, \lambda_\text{p}^\text{bg}, \lambda_\text{g}, \lambda_\text{i} ,\lambda_\varepsilon, \lambda_s$) are set to $(2,1,0.66,2,4,10^5,3)$ in the first stage, $(1,2,1,2,4,10^5,2)$ in the second stage.

We perform PTI~\cite{pti} optimization in the third stage for 500 iterations. We multiply $L_\text{LPIPS}$ with 2 and use the default PTI parameters.

For PTI in the EG3D projection, we follow EG3D's optimization procedure, which runs 500 iterations for $\W$ latent optimization and 500 iterations for tuning.


\subsection{Running Time Comparison}
We measured our runtime on a single GPU NVIDIA RTX 2080Ti with AMD Threadripper 2920x. We used around 21 minutes per input pair. The construction time for the guide images is around 13 minutes: 5 minutes for EG3D projection and 6–10 minutes for EG3D warping. Multi-view latent optimization requires around 8 minutes.

Note that we have not optimized our code, and many of the 3D pre-processing steps (8 mins), such as our occlusion test and marching cube, can be implemented on the GPU with real-time speed (currently, it's in python). Techniques such as PSP~\cite{richardson2021encoding} can speed up and perform our EG3D projection with a single network inference. The rest of the pipeline takes about 8 mins, which is in the same order as LOHO: 15 mins, Barbershop: 5 mins, and StyleYourHair: 8 mins. HairNet still does require StyleGAN projections and PTI~\cite{pti} for pre-precessing, which take several minutes per image.

\section{Construction of Our FFHQ-S Testset}
\label{apx:create_dataset}
This section explains the criteria used for determining the four scenarios in 12-Config FFHQ-S in Table~\ref{table:user_study}. All criteria are computed from raw $I_f$ and $I_h$. We skip any pair in which the number of hair pixels in $I_h$ is less than 5 percent of all the pixels in the image.
\label{sec:metric}

\indent\textbf{Pose Misalignment:} The criterion for this scenario is based on the difference between the yaw angles of $I_f$ and $I_h$, estimated from facial keypoints~\cite{yawgithub}. The angle difference between [0,15) is indicated with `-', [15, 30) with a checkmark, and [30, 45) with double checkmarks in Table~\ref{table:user_study}.

\indent\textbf{Needs Face Inpainting:} 
This criterion tests whether $I_f$'s face is occluded, by checking if the number of pixels in $H_\text{face} - F_\text{face}^\text{k}$ is greater than 10 percent of all pixels.

\indent\textbf{Needs BG Inpainting:} 
This criterion tests whether a substantial number of background pixels need to be hallucinated. This happens when $I_f$'s hair is smaller than $I_h$'s hair or, specifically, when the number of pixels of $H_\text{hair}$ - $F_\text{hair}$ is greater than 15 percent of all pixels.

\indent\textbf{\bm{$I_h$} Contains Hat:} This criterion tests whether some part of the reference hair in $I_h$ is missing due to hat wearing by checking if the number of pixels in $H_\text{hat}$ is greater than 5 percent of all pixels.

\section{Additional Experiments}
\label{apx:other_exp}
\subsection{Can Prior Work Solve Challenging Scenarios With a Good Target Segmentation Mask Constructed Using Our Rules?
} \label{sec:barber_our_mask}

In this section, we construct a target segmentation mask based on our rules in Section \ref{sec:met_align} / Appendix \ref{sec:guide} and use it in place of the original mask used in StyleYourHair~\cite{kim2022styleyourhair} or Barbershop~\cite{barbershop}, then compare their results with ours. 

\begin{figure}[t]
\centering
\includegraphics[width=1\linewidth\vspace*{-0.4em}]{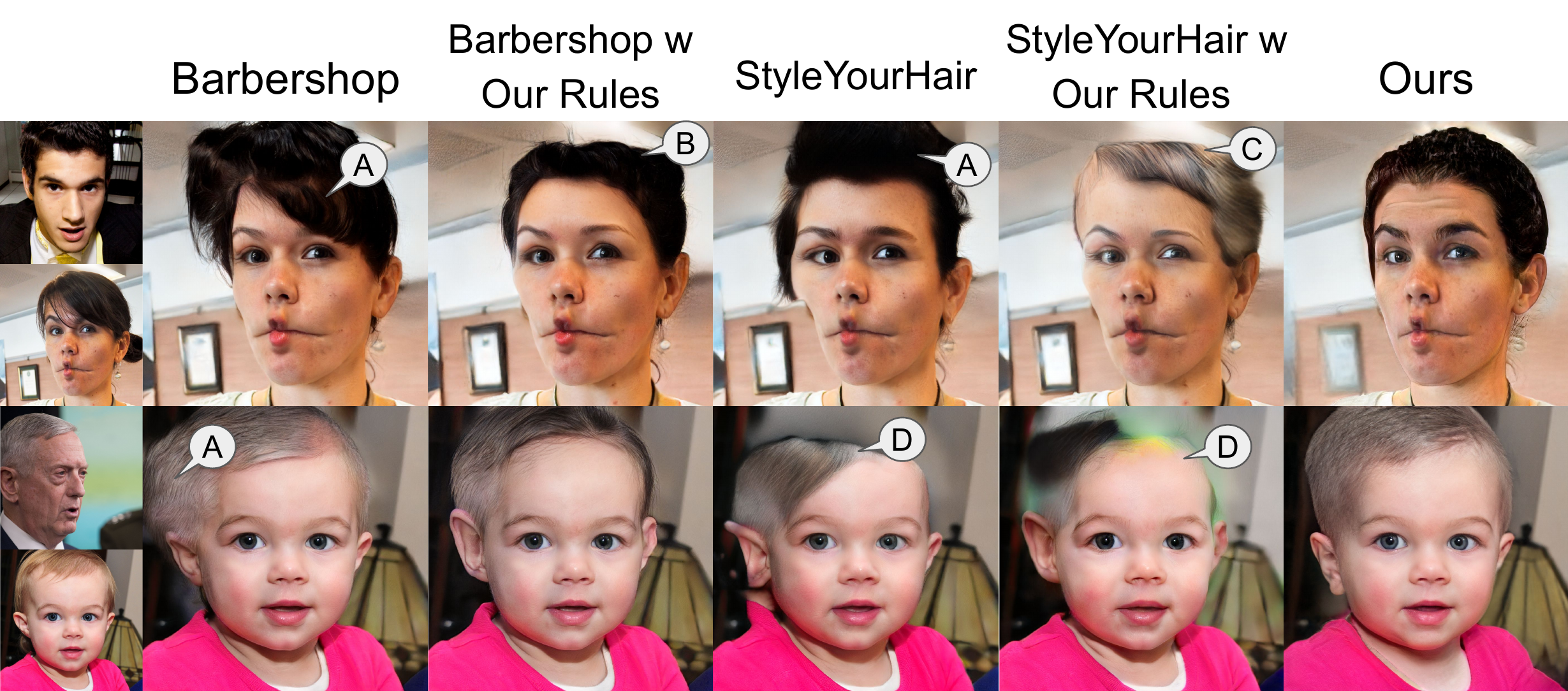}
  \caption{Results when StyleYourHair~\cite{kim2022styleyourhair} and Barbershop~\cite{barbershop} use our rules to create the target segmentation mask: Bangs and unnecessary hair are totally removed (A). However, the modified target segmentation is not realistic (similar to our guide) and produces unrealistic hair shapes (B) with poorer color reproduction (C). Although our modified target segmentation can improve Barbershop's result in the second row, it cannot improve StyleYourHair, whose technique also includes a hair-warping stage (D).
  }
\label{fig:barber_our_input}
\end{figure}

Figure~\ref{fig:barber_our_input}(A) shows that the original Barbershop and StyleYourHair fail to completely remove bangs from the forehead or add more hair that makes the hairstyle incorrect, but the modified version using our provided target segmentation mask can remove the bangs completely as well as any unnecessary hair. Compared to our method, this modified version still produces (B) unnatural hairstyles with (C) poorer color reproduction.

Unlike Barbershop and StyleYourHair, which use a high-dimensional latent space that can overfit the error-prone target segmentation mask, our method can better refine the boundaries between semantic regions by first predicting the output in original latent space that ensures natural-looking hair before refining the output in the extended space with LPIPS pre-masking for seamless blending.
Importantly, this shows that our state-of-the-art quality requires not only our well-designed guide image but also our multi-view latent optimization that uses the guide image in a flexible and effective manner.

\subsection{Quality of Hairstyle Transfer}
\label{sec:soa_comp_transferfid}
Following StyleYourHair~\cite{kim2022styleyourhair}, we compute the FID score to compare the distributions of the results and real images.
However, note that FID is not ideal for this task because an algorithm that minimally changes or does not change any hair at all can achieve the best performance.
We use the same dataset in Section \ref{sec:soa_comp_face}.
All methods yield comparable results and are roughly equivalent, shown in Table~\ref{table:fid_hair}. LOHO, which has face and background blending, receives the best FID score of 20.7. However, according to our user study, people are less likely to prefer LOHO over other methods. 


\begin{table}[t]
\begin{center}
\setlength{\tabcolsep}{0.75em}
\resizebox{\columnwidth}{!}{
\begin{tabular}{l|c|c|c|c}
\toprule

       &LOHO &Barbershop&StyleYourHair&Ours \\ \midrule
Hairstyle (FID $\downarrow$) & \textbf{20.72} & 21.22 & 21.64 & \underline{21.02} \\ \bottomrule
\end{tabular}}
\caption{ \label{table:fid_hair} FID scores after performing the hairstyle transfer (Section \ref{sec:soa_comp_transferfid}).
}
\end{center}
\end{table}

\subsection{Comparison of the ability to preserve details.}
\begin{figure}[t]
\centering
\includegraphics[width=1\linewidth\vspace*{-0.2em}]{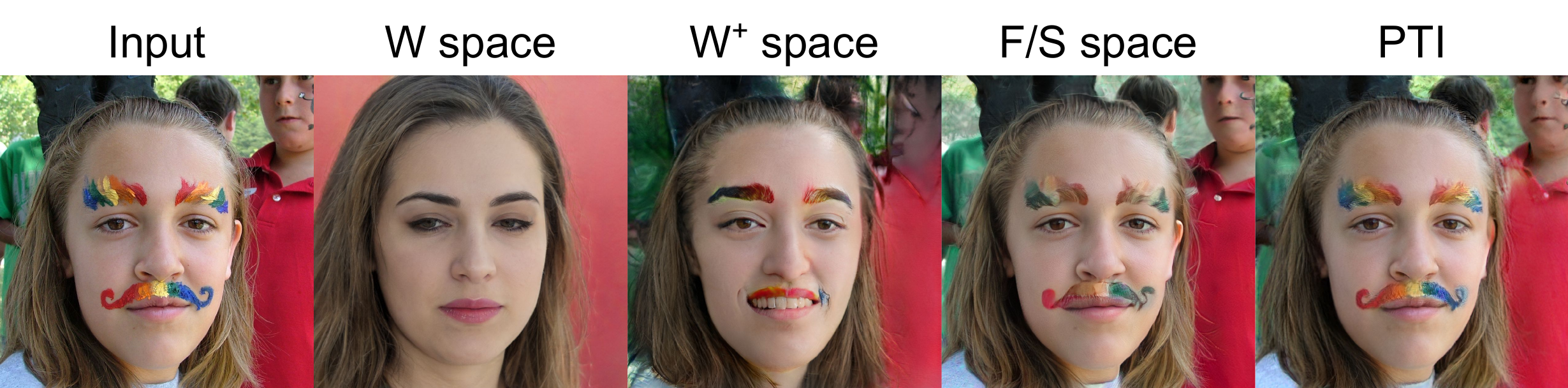}
  \caption{
  Comparison of different reconstruction techniques ordered by the degree of freedom. PTI with the highest degree of freedom successfully reconstructs the texture and color details of a woman with heavy makeup, as shown in this example.
  }
\label{fig:detail}
\end{figure}

To evaluate the effectiveness of each method in preserving image details, we perform an image reconstruction task without employing the hairstyle transfer technique, in order to eliminate any potential external factors.
Figure~\ref{fig:detail} demonstrates that the ability to preserve details highly depends on the degree of freedom (sorted descendingly): pivot tuning inversion (PTI), $\FS$ space, $\WP$ space, and $\W$ space.

\section{Additional Ablation Studies}
\label{apx:other_abla}
\subsection{Qualitative Ablation Studies}
Ablation studies with quantitative metrics are highly difficult to do because there is no ground truth and the existing metrics such as FID score are unreliable. For example, dropping the latent sharing (Figure ~\ref{fig:ablation}a) can produce a better FID score despite the clearly wrong hairstyle, which is not captured by FID. 
Nonetheless, we identified three most crucial components and conducted an additional user study (30 randomly sampled input pairs, each evaluated by 3 different users). The users preferred our full method 27.9\% of the time, compared to Config i) 23.4\%, ii) 4.5\%, iii) 25.2\% in Figure~\ref{fig:ablation}.

\subsection{Ablation studies on latent sharing and the loss function.}
\begin{figure*}[t]
\centering
\includegraphics[width=1\linewidth\vspace*{-0.4em}]{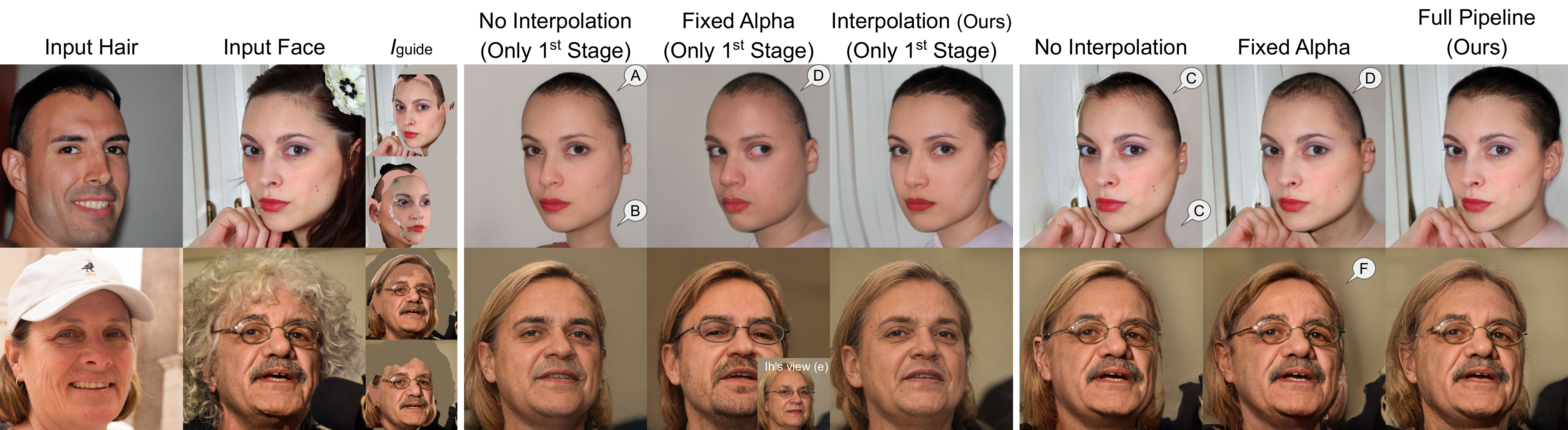}
  \caption{Ablation study on $\W$ space latent optimization. Without our latent interpolation technique, the results are less realistic (A,B,C,D).
  If we fix the interpolation coefficient $\alpha$, some information, such as color, may not be shared between two viewpoints (E), resulting in poorer background details (F). }
\label{fig:more_ablation_latent}
\vspace{-0.5cm}
\end{figure*}

\begin{figure}[t]
\centering
\includegraphics[width=1\linewidth\vspace*{-0.2em}]{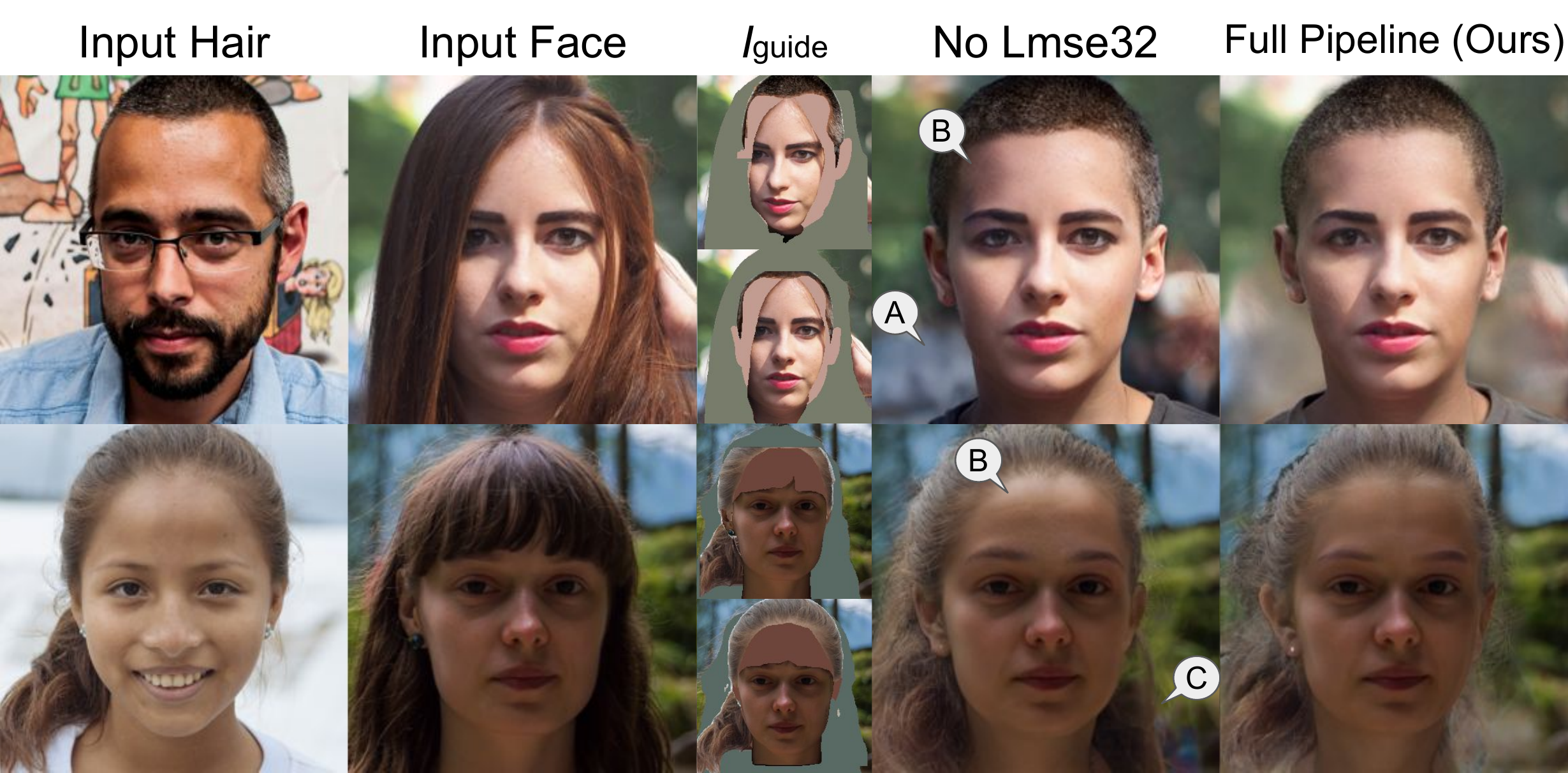}
\caption{Ablation study on $L_\text{MSE}^{32}$. Without this loss, the regions not constrained by any loss function can be hallucinated freely.
While this can lead to some positive results, such as better background details (A), it can also produce unrealistic shading (B) or excessive hair (C).}
\label{fig:more_ablation_small}
\end{figure}

We test additional ablation configurations on our multi-view $\W$ space latent optimization: i) sharing $w$ latent code without interpolation, and ii) fixing $\alpha$ to 0.5. The results are shown in Figure~\ref{fig:more_ablation_latent}. We also test our complete pipeline iii) without using $L_\text{MSE}^{32}$ and show the results in Figure~\ref{fig:more_ablation_small}.  

Instead of sharing the latent code with our interpolation technique, we optimize a shared latent code, which is fed to the last $l$ layers of StyleGAN ($w_{l:18}$) for optimizing both views (Config i). This is similar to the concept of sharing $w^+$ latent code in Section \ref{sec:met_latshare}, but with fewer parameters (19x512 fewer than $w^+$ code, and 512 more than $w$ code). We add the latent similarity loss (see Section \ref{sec:met_loss}) between the latent codes used for StyleGAN's early layers and the new latent code to ensure that all latent codes are similar, which can avoid overfitting. 

In Figure \ref{fig:more_ablation_latent}, Config i) may produce unrealistic head shapes (Figure \ref{fig:more_ablation_latent}-A) or necks (Figure \ref{fig:more_ablation_latent}-B in the first stage. This artifact still manifests in the second and third stages of optimization (Figure \ref{fig:more_ablation_latent}-C. This Config i) has a higher degrees of freedom and thus can fit unrealistic guide images.

When we fix the interpolation coefficient $\alpha$ (Config ii), the results also contain unrealistic head shapes (Figure \ref{fig:more_ablation_latent}-D), and the colors of the face, hair, or background may look different in each viewpoint (Figure \ref{fig:more_ablation_latent}-E). As a result, the hallucinated background from this configuration becomes less accurate (Figure \ref{fig:more_ablation_latent}-F). Our proposed method helps alleviate this issue by forcing the shared part to be similar via random interpolation. In particular, the optimizer is encouraged to use the same values for both latent codes so that their interpolation with any $\alpha$ will remain stationary.

Without $L_\text{MSE}^{32}$ (Config iii), the optimization in each stage would not try to reproduce the overall appearance of $\iguide$ and is free to synthesize arbitrary content on regions not constrained by any loss function. This can result in more realistic background details (Figure~\ref{fig:more_ablation_small}-A) but less realistic shading (Figure~\ref{fig:more_ablation_small}-B) or excessive hair (Figure~\ref{fig:more_ablation_small}-C).

\section{Comparison to concurrent work, HairNet} \label{apx:hairnet}
\begin{figure}[t]
\centering
\includegraphics[width=1\linewidth\vspace*{-0.2em}]{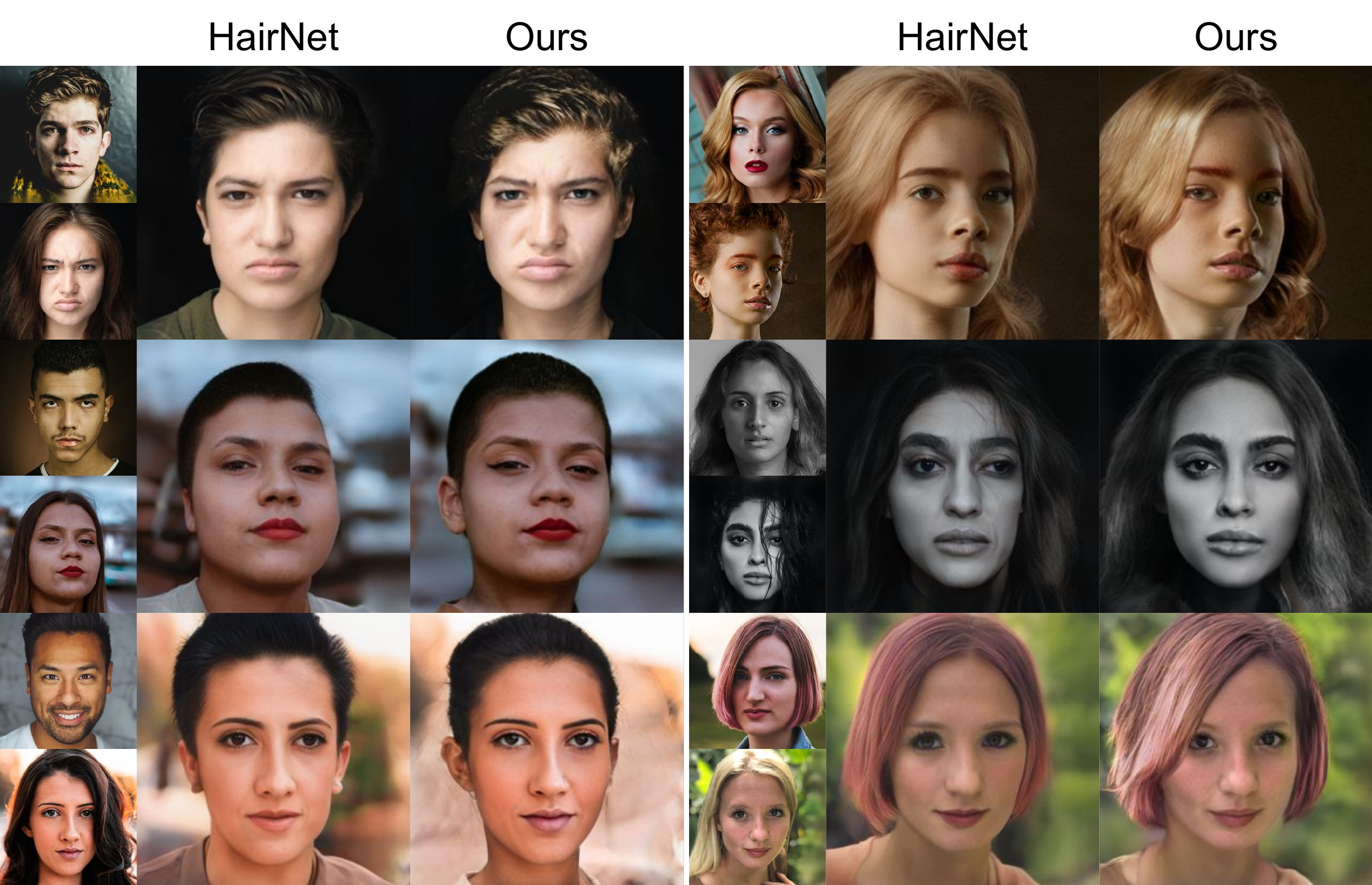}
\caption{Comparison to concurrent work, HairNet. Our method is better at preserving input face identity and input hair details (top row), hair rotation (left side, 2nd row), and face inpainting (right side, 2nd row). Conversely, HairNet is better at background inpainting (left side, last row) and producing realistic hair and lighting details (right side, last row).
}
\label{fig:hairnet}
\end{figure}

HairNet is also capable of pose-invariant hairstyle transfer. Unfortunately, their official code is not publicly available during our study. We provide a qualitative comparison in Figure~\ref{fig:hairnet} and conducted a user study on their selected input pairs (380 input pairs, each evaluated by 3 different users). 
The participants preferred our results 52.4\% of the time, whereas HairNet was selected for 47.6\%.

We observe that HairNet often fails to preserve input face identity or hair details (Figure~\ref{fig:hairnet}, top row). Our method excels at rotating the input hair (left side, 2nd row) and restoring unseen facial features (2nd row, right side). Conversely, HairNet is better at filling in background details (left side, last row) and producing realistic hair and lighting details (right side, last row).

In addition, we calculate the maximum, minimum, and average pose difference of HairNet's input pairs, which are 22.8, 0.0, and 4.7, respectively. We suggest using our datasets for further analysis of the results in future work.


\section{Additional Results}
\label{apx:other_results}
\begin{figure*}[t]
\begin{center}
\includegraphics[width=1\textwidth\vspace*{-0.5em}]{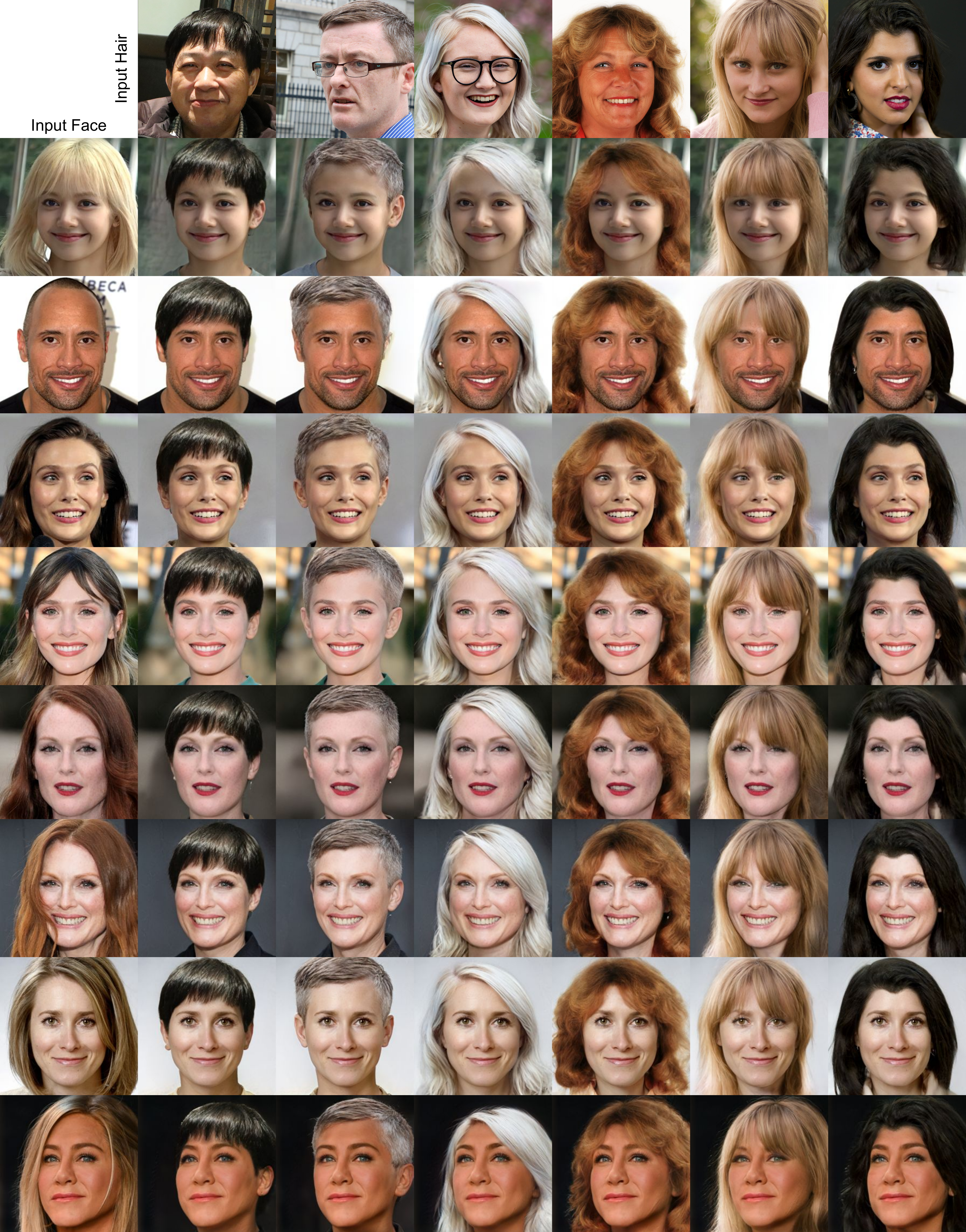}
\end{center}
  \caption{
  Our method can transfer the hairstyle from any reference hair image in the top row to an input person~\cite{lisa,the_rock,scarlet_witch,olsen,Julianne,Julianne2,Kaja,jennifer} in each row.
  }
\label{fig:result_ours_real}
\end{figure*}

\begin{figure*}[t]
\begin{center}
\includegraphics[width=1\textwidth\vspace*{-0.5em}]{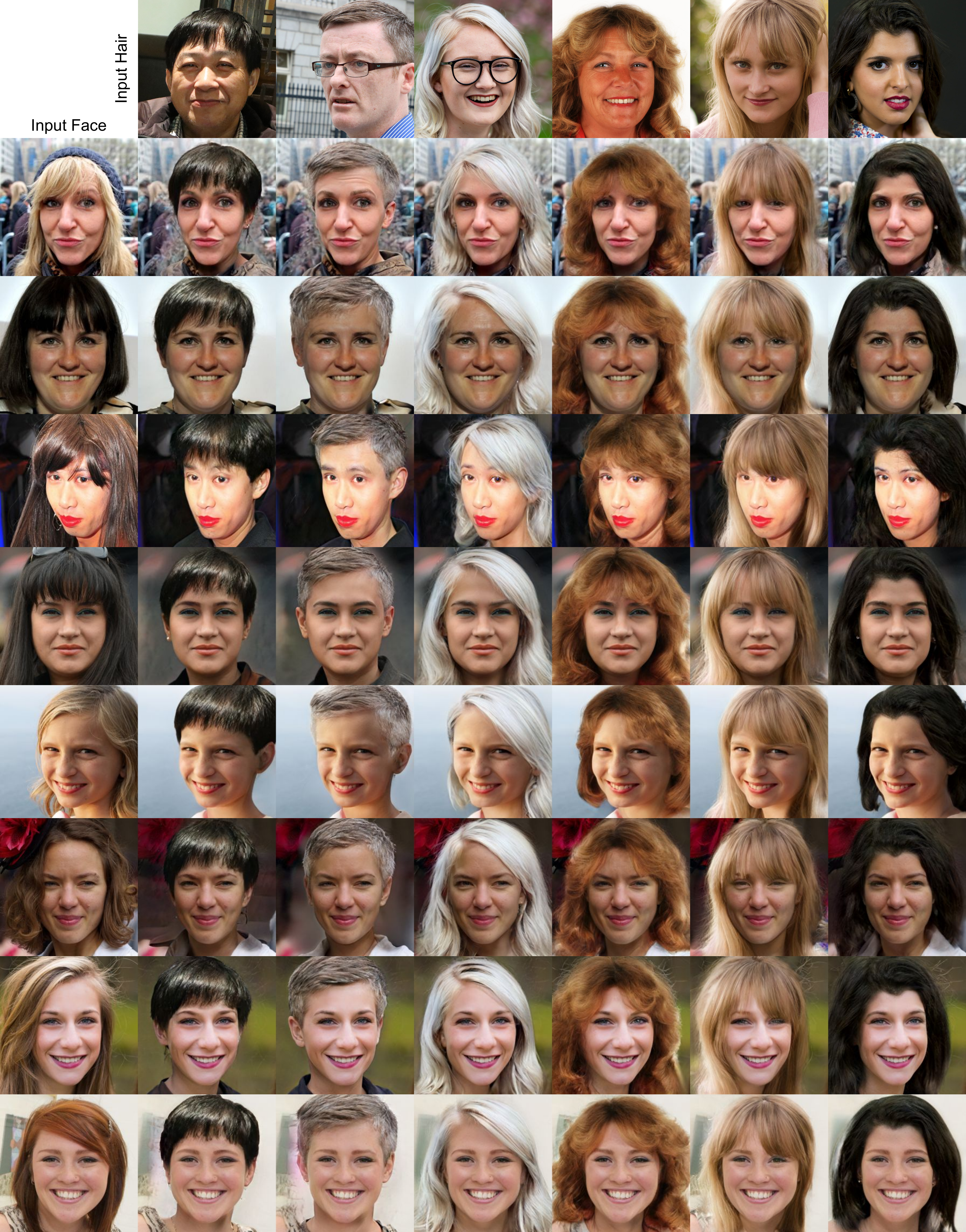}
\end{center}
  \caption{
  Our method can transfer the hairstyle from any reference hair image in the top row to an input person in each row.
  }
\label{fig:result_ours}
\end{figure*}

\begin{figure*}[t]
\begin{center}
\includegraphics[width=1\textwidth\vspace*{-0.5em}]{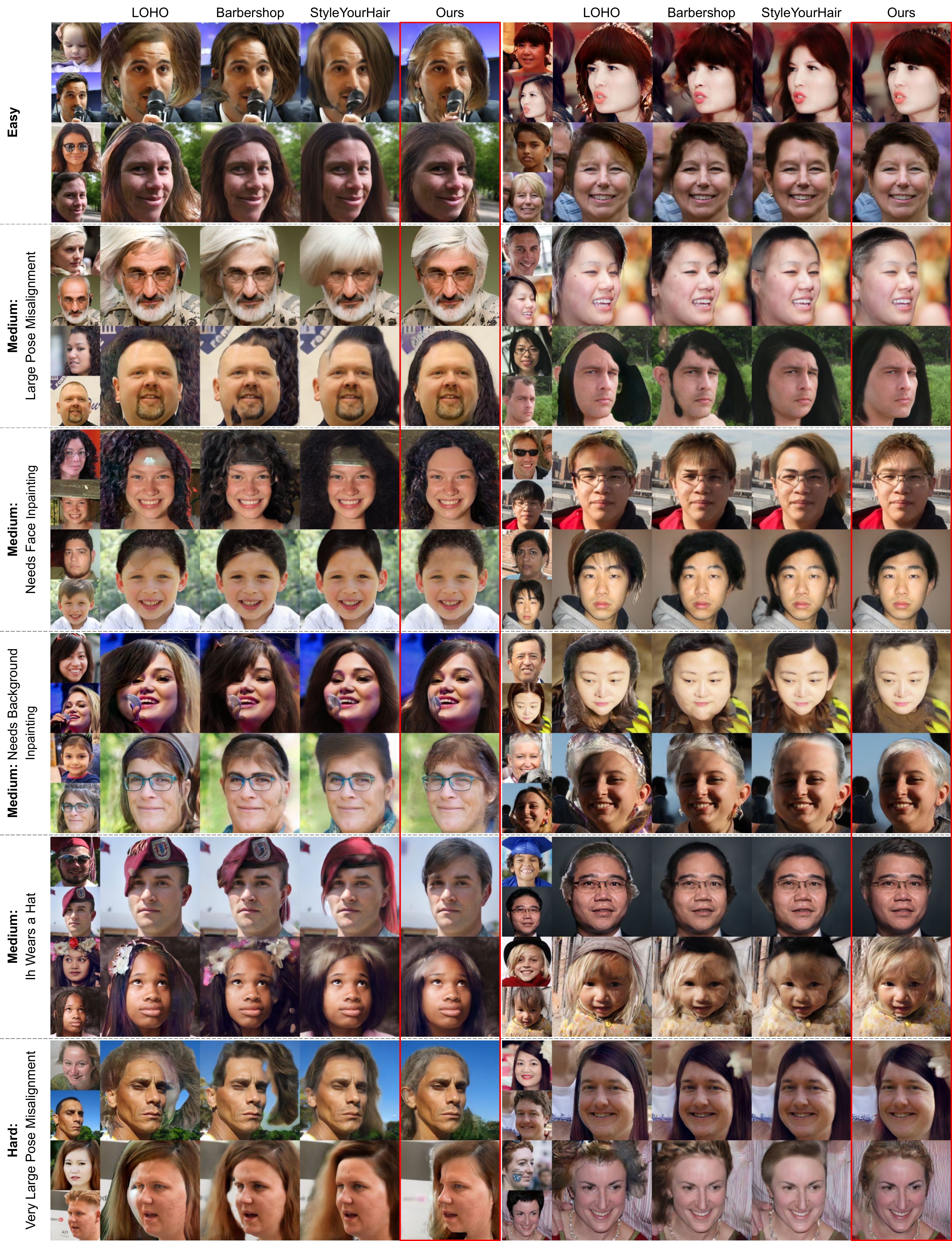}
\end{center}
\vspace{-0.2cm}
 \caption{
  Random test samples from FFHQ-S for comparison to StyleYourHair~\cite{kim2022styleyourhair}, Barbershop \cite{barbershop}, and LOHO~\cite{saha2021LOHO}.
  }
\label{fig:result1}

\end{figure*}

\begin{figure*}[t]
\begin{center}
\includegraphics[width=1\textwidth\vspace*{-0.5em}]{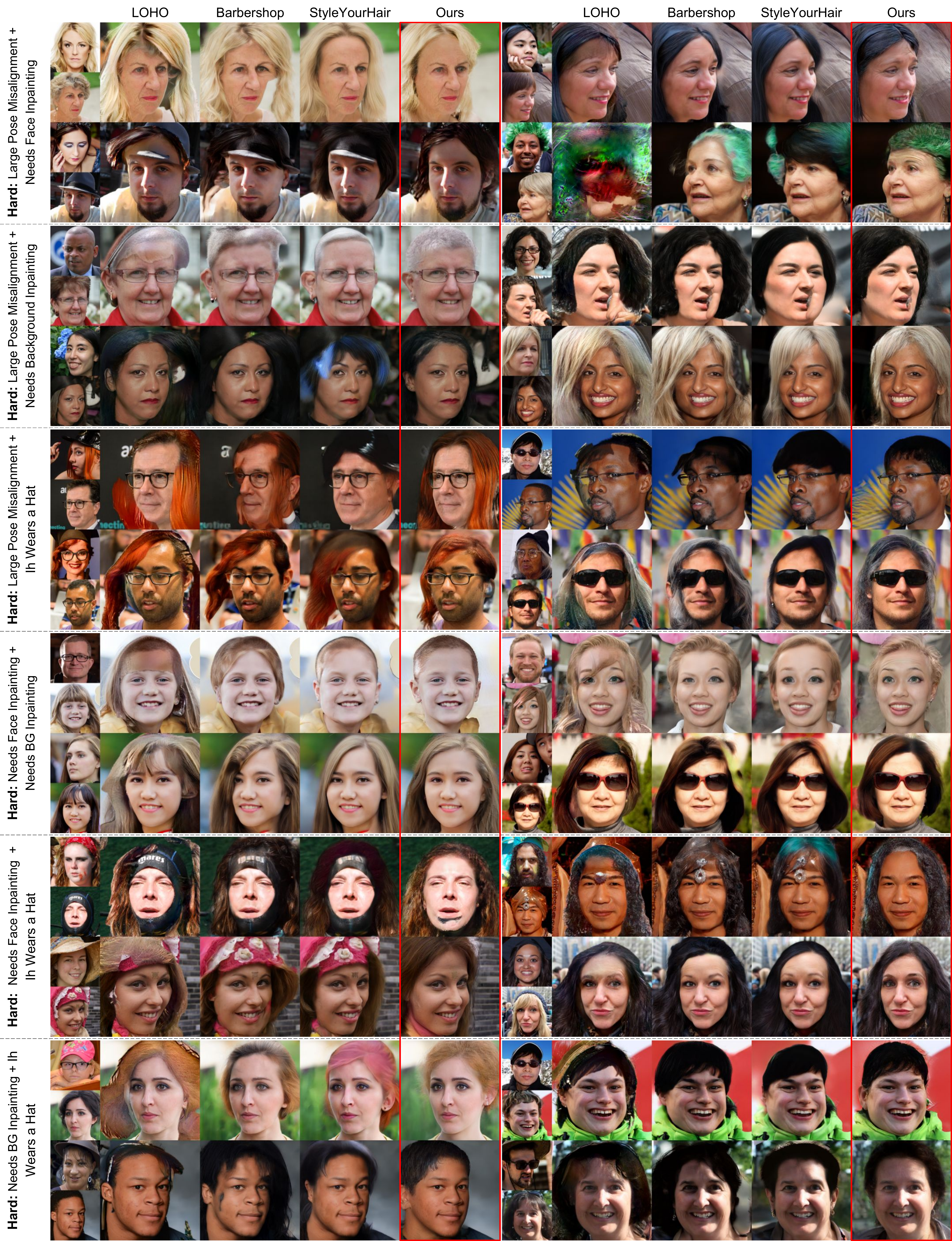}
\end{center}
\vspace{-0.2cm}
\caption{
  Random test samples from FFHQ-S for comparison to StyleYourHair~\cite{kim2022styleyourhair}, Barbershop \cite{barbershop}, and LOHO~\cite{saha2021LOHO}.
  }
\label{fig:result2}
\end{figure*}

\begin{figure*}[t]
\begin{center}
\includegraphics[width=1\textwidth\vspace*{-0.5em}]{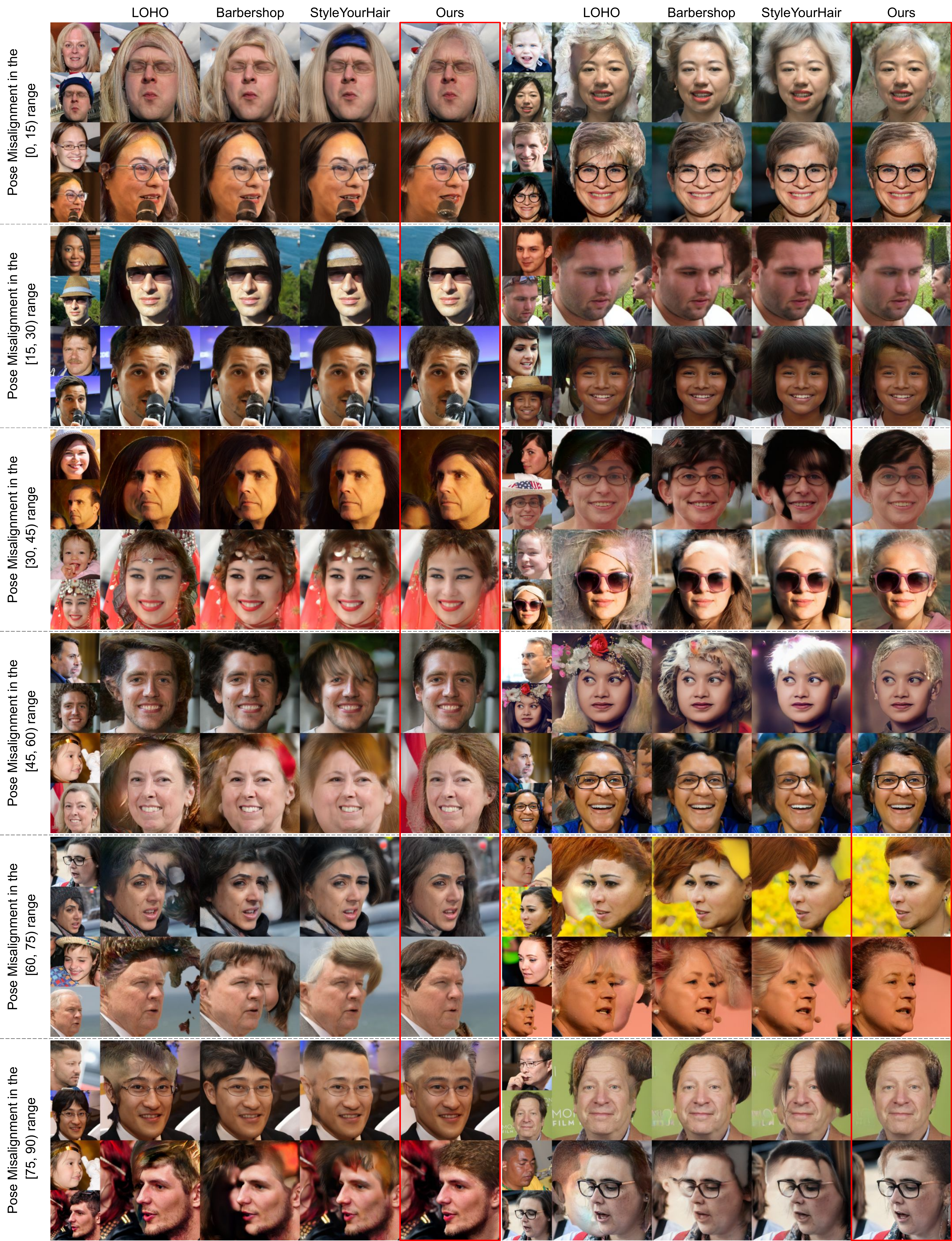}
\end{center}
\vspace{-0.2cm}
\caption{
  Random test samples from FFHQ-P for comparison to StyleYourHair~\cite{kim2022styleyourhair}, Barbershop \cite{barbershop}, and LOHO~\cite{saha2021LOHO}.
  }
\label{fig:result3}
\end{figure*}

In this section, we present more qualitative results in Figure \ref{fig:result_ours_real}, \ref{fig:result_ours}, and random test samples from our user study in Figure \ref{fig:result1}, \ref{fig:result2}, and \ref{fig:result3}.



\section{Failure Cases}
\label{apx:fail}
Figure \ref{fig:fail} compiles a set of our failure cases.
We cannot transfer hairstyles with incorrect semantic regions (Figure~\ref{fig:fail}-A). Large errors from the keypoint detector can place the hair in the wrong place (Figure~\ref{fig:fail}-B).
Poor EG3D projection results may make the hair look different from the reference hair (Figure~\ref{fig:fail}-C).
Some hair may be blended into the background if the reference hair has a similar color as the background (Figure~\ref{fig:fail}-D).
Other failure cases include mismatched lighting conditions (Figure~\ref{fig:fail}-E) and highly unusual hairstyles (Figure~\ref{fig:fail}-F).


\section{Potential Negative Societal Impact}
\label{apx:negative}
Even though the hairstyle transfer results from our method are realistic, they are considered \emph{fake} images and can have similar uses and misuses as DeepFake. The results may contain some artifacts that another network can easily detect~\cite{dong2022think}. Our method relies on multiple pretrained networks, which may contain race, gender biases. Our method also may not work as well on people who are less represented in the training set.

We also consider the validity of the copyright for the reference hair after it has been transferred to our input, as well as the possibility of transferring the hairstyles of others without their permission.
To circumvent this, we suggest that the use of both input face and reference hair be under Creative Commons licenses to prevent any potential conflicts.


\ifarxiv \clearpage  \fi

\end{document}